\documentclass{article}

\usepackage[final]{corl_2022} 

\usepackage[pdftex]{graphicx}
\usepackage{amsmath}
\usepackage{amsfonts}
\usepackage{bm}
\usepackage[linesnumbered,ruled,vlined]{algorithm2e}
\usepackage[utf8]{inputenc} 
\usepackage[T1]{fontenc}   
\usepackage{hyperref}      
\usepackage{url}           
\usepackage{booktabs}      
\usepackage{amsfonts}      
\usepackage{nicefrac}      
\usepackage{microtype}     
\usepackage{xcolor}

\usepackage{wrapfig}
\usepackage{subcaption}
\usepackage{afterpage}
\usepackage{tikz}
\usepackage{pgfplots}
\pgfplotsset{compat=1.17}

\usetikzlibrary{positioning} 
\usetikzlibrary{calc}
\usetikzlibrary{pgfplots.groupplots}

\SetKwInput{KwInput}{Input}                
\SetKwInput{KwOutput}{Output}              
\newcommand{\w}{\bm{w}}
\newcommand{\ctx}{\bm{c}}

\newcommand{\z}{\bm{z}}
\newcommand{\obs}{\bm{o}}
\newcommand{\Obs}{\bm{O}}
\newcommand{\bphi}{\bm{\phi}}
\newcommand{\btau}{\mathbf{\tau}}

\DeclareMathOperator*{\argmin}{arg\,min}
\newcommand{\E}[2]{\mathbb{E}_{#1}\left[ #2\right]}

\usepackage[acronym,nohypertypes={acronym,notation}]{glossaries}
\newacronym{il}{IL}{Imitation Learning}
\newacronym{gail}{GAIL}{Generative Adversarial Imitation Learning}
\newacronym{gmm}{GMM}{Gaussian Mixture Model}
\newacronym{promp}{ProMP}{Probabilistic Motion Primitive}
\newacronym{mp}{MP}{Movement Primitive}
\newacronym{drex}{D-REX}{Disturbance-based Reward Extrapolation}
\newacronym{em}{EM}{Expectation Maximization}
\newacronym{vips}{VIPS}{Variational Inference by Policy Search}
\newacronym{eim}{EIM}{Expected Information Maximization}
\newacronym{bc}{BC}{Behavioral Cloning}
\newacronym{ppo}{PPO}{Proximal Policy Optimization}
\newacronym{dof}{DoF}{Degrees of Freedom}
\newacronym{rl}{RL}{Reinforcement Learning}
\newacronym{irl}{IRL}{Inverse Reinforcement Learning}

\newacronym{mdn}{MDN}{Mixture Density Network}
\newacronym{vigor}{VIGOR}{Versatile Imitation from Geometrically Observed Representations}

\title{Inferring Versatile Behavior from Demonstrations by Matching Geometric Descriptors}

\author{%

  Niklas Freymuth$^1$\thanks{correspondence to \texttt{niklas.freymuth@kit.edu}} \\
 \And
   Nicolas Schreiber$^1$ \\
 \And
 Philipp Becker$^1$
 \And 
    Aleksandar Taranovic$^{1,2}$\\
\And
   Gerhard Neumann$^1$ \\
\AND 
     $^1$Autonomous Learning Robots\\
   Karlsruhe Institute of Technology\\
   Karlsruhe, Germany
\And 
    $^2$Bosch Center for Artificial Intelligence (BCAI)\\
Renningen, Germany
}

\begin{document}
\maketitle


\begin{abstract}
Humans intuitively solve tasks in versatile ways, varying their behavior in terms of trajectory-based planning and for individual steps.
Thus, they can easily generalize and adapt to new and changing environments.
Current Imitation Learning algorithms often only consider unimodal expert demonstrations and act in a state-action-based setting, making it difficult for them to imitate human behavior in case of versatile demonstrations.
Instead, we combine a mixture of movement primitives with a distribution matching objective to learn versatile behaviors that match the expert's behavior and versatility. 
To facilitate generalization to novel task configurations, we do not directly match the agent's and expert's trajectory distributions but rather work with concise geometric descriptors which generalize well to unseen task configurations.
We empirically validate our method on various robot tasks using versatile human demonstrations and compare to imitation learning algorithms in a state-action setting as well as a trajectory-based setting. 
We find that the geometric descriptors greatly help in generalizing to new task configurations and that combining them with our distribution-matching objective is crucial for representing and reproducing versatile behavior.
\end{abstract}


\section{Introduction}
\label{sec:introduction}

\begin{figure}
    \centering
    \resizebox{\textwidth}{!}{\input{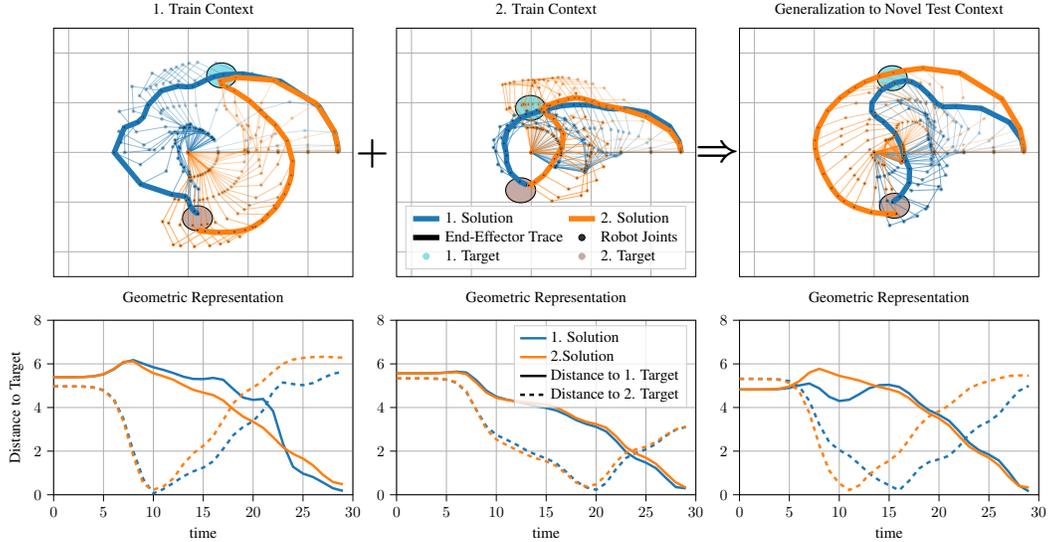}}
    \caption{End-effector traces (Top) and distances to target centers (Bottom) for a planar point-reaching task on $3$ different contexts. (Left, Middle) Human experts solve the task by reaching the blue target and ending their trajectory in the grey one. (Right) \gls{vigor} matches distributions of behavioral descriptors, such as the depicted target distances, of these demonstrations. This produces versatile behavior in unseen contexts, such as new target positions. Depicted are $2$ different component means from a trained \gls{gmm} Policy. For each context, we show the joint configurations for the first trajectory in blue and the second trajectory in orange.}
    \label{fig:figure_one}
\end{figure}

\gls{il}~\citep{schaal1996learning, argall2009survey, hussein2017imitation} from human demonstrations is challenging as humans often solve tasks in versatile ways. 
Even the same person might solve a task differently when confronted with it multiple times.
Behavior can be versatile in terms of task-level decisions, such as planning a route to a target, and individual actions, such as randomly pausing during a movement to think about the next steps.
Most recent \gls{il} approaches~\citep{fu2018learning, ho2016generative, brown2019extrapolating, brown2020better} model the behavior in state-action space using Gaussian policies, which assumes the behavior is unimodal and cannot capture this planning versatility. 
Yet, this assumption is violated by most human expert datasets, often causing poor generalization
to human demonstrations~\citep{orsini2021matters}.
Additionally, 
\gls{il} approaches often need immense amounts of data to achieve generalization \cite{osa2018algorithmic, orsini2021matters}. 
We tackle these challenges by a feature matching approach to \gls{il} that is able to generalize to novel contexts from a small amount of expert demonstrations.
Such contexts are (typically low-dimensional) vectors that describe a specific task configuration within a family of related tasks, such as e.g., the coordinates of goal locations to reach.
The resulting approach, \gls{vigor}, models distributions that match expert trajectories in terms of concise geometric behavioral descriptors.
\gls{vigor} represents distributions using \glspl{gmm} over \glspl{promp}~\citep{paraschos2013probabilistic}, 
and utilizes descriptors that are designed to abstract away from concrete contexts and thus facilitate generalization to novel contexts in a sample-efficient way.
An example of such descriptors would be geometric features in the form of (only) the distance between a target and the robot's end-effector, plus features that denote the smoothness of the robot.
While similar behavioral descriptors have found use in Inverse Optimal Control \citep{englert2017inverse, englert2018learning} to recover cost functions from demonstrations, common \gls{il} approaches instead use state representations that include both relative and absolute information~\citep{ho2016generative, mandlekar2021matters}.
The exact form of this set of behavioral descriptors is task-dependent and allows the user to include domain knowledge.
Figure \ref{fig:figure_one} shows how \gls{vigor} works on an exemplary reaching task. 
For training the \glspl{gmm}, we rely on recent approaches~\citep{arenz2018efficient, arenz2020trust, Becker2020Expected} for distribution matching and Variational Inference.
For inference, we train individual \glspl{gmm} for each training context and an arbitrary number of
test contexts.

We evaluate \gls{vigor} on a suite of versatile robotic tasks, where demonstrations are collected from human experts through teleoperation.
We compare our method to standard \gls{bc}~\citep{bain1995framework}, \gls{bc} with a \gls{gmm} policy~\citep{mandlekar2021matters, zhou2020movement}, \gls{gail}~\citep{ho2016generative} and \gls{irl} techniques~\citep{brown2020better}. 
We find that these methods either fail to learn useful behavior because they average over multiple solutions, or that they are unable to capture the full versatility of the demonstrations.
Contrary to this, \gls{vigor} accurately models highly versatile behavior from already a small number of trajectories demonstrated by human experts.
We perform extensive ablation studies to showcase the importance of different parameters and design choices.
Datasets and code can be found at \url{https://www.github.com/NiklasFreymuth/VIGOR}.

To summarize, our list of contributions is as follows:
\begin{itemize}
    \item We infer \textit{multi-modal distributions} of desired trajectories from highly versatile human expert trajectories by matching distributions over behavioral descriptors between learner and expert.
    \item Our approach is, to the best of our knowledge, the first Adversarial Imitation Learning method to utilize \textit{concise behavioral descriptors} to facilitate generalization to novel contexts from already a small number of demonstrations.
    \item We conduct \textit{experiments with human demonstrations} in simulation and on a real robot and find that \gls{vigor} accurately models highly versatile human behavior, outperforming various Imitation Learning baselines.
\end{itemize}


\section{Related Work} 

\textbf{Imitation Learning for Skills.}
A common way to represent skills over trajectories is via \glspl{mp}~\citep{schaal2006dynamic, paraschos2013probabilistic}. 
While learning individual \glspl{mp} from human demonstrations is a simple regression problem, modeling versatile behavior requires a more sophisticated model that can represent such multi-modality, e.g., a mixture model.
Using \gls{em}~\citep{dempster1977maximum}, both~\citep{mulling2013learning} and~\citep{ewerton2015learning} fit a single GMM over \gls{mp} parameters of multiple demonstrations for different contexts and use them to generalize to novel contexts by conditioning. 
Yet, as they only fit a single GMM for all contexts, they do not explicitly focus on representing versatility.
Recent work~\citep{pervez2018learning} instead learns a mixture over contextualized \glspl{mp} using Gaussian Mixture Regression. 
This has been extended to non-linear relations between MP parameters and context
\citep{zhou2020movement} by using \glspl{mdn}~\citep{bishop1994mdn}. 
While the above approaches work for tasks with a small number of modes, optimizing \glspl{mdn} can be challenging in the case of many modes in the demonstrations.
\citet{osa2017guiding} use planning algorithms to reproduce human demonstrations in a trajectory-based setting. 
They tune the parameters of a cost function such that the planned trajectories match the demonstrations and use the extracted cost function to guide the trajectory optimization process. 
Yet, this approach is limited to uni-modal demonstrations and requires engineered cost functions. 

\textbf{Imitation Learning by Distribution Matching.}
Classical \gls{il} approaches~\citep{abbeel2004apprenticeship, ziebart2008maximum} employed feature expectation matching to learn a policy from behavioral descriptors of expert demonstrations.
Similarly, a recent body of work \citep{englert2017inverse, englert2018learning} in Inverse Optimal Control utilizes geometric behavioral descriptors that are similar to ours to learn cost functions for manipulation tasks from a few demonstrations.
Whereas these approaches linearly match the moments/expectations of their features, our method instead matches a versatile distribution over non-linear features using the reverse Kullback-Leibler Divergence (KL)~\citep{kullback1951information}. Directly matching the distribution rather than its moments means that our method can represent complex and multi-modal distributions.
More recently, a new class of distribution matching approaches has gained popularity with the advent of Generative Adversarial Nets~\citep{goodfellow2014generative}. Starting from \gls{gail}~\citep{ho2016generative}, a whole class of distribution matching based \gls{il} approaches was developed~\citep{fu2018learning, kostrikov2018discriminator, torabi2019adversarial, kostrikov2019imitation, ghasemipour2020divergence, zhang2020f}.
All of these approaches commonly work in a step-based setting and minimize some divergence between the state-action occupancy or state marginals.
They generally do not consider versatile behavior, and additionally need to interact with the environment during training in order to generate states to match. 
We instead work in a trajectory-based setting , which allows us to abstract away the environment's dynamics and learn entirely offline, i.e., without any environment interactions.
Common methods in this setting usually employ a maximum likelihood approach, which, as previously discussed, performs poorly on versatile data. 
As a solution to this problem,
\citet{Becker2020Expected} propose using the Information-Projection~\citep{murphy2012machine} instead, which is able to focus on individual modes of data rather than averaging over all modes of data.
This approach has been extended to \gls{irl}~\citep{freymuth2021versatile}.
Here, we build on these methods, generalizing them to sequential data and geometric behavioral descriptors.

\textbf{Comparison-Based Approaches.}
\citet{brown2019extrapolating} perform \gls{irl} by utilizing provided rankings of demonstrations to train a discriminator on a comparison-based loss.
As the discriminator learns to favor samples with a high rank, it can be used as a reward function after training.
This has been extended by \gls{drex}~\citep{brown2020better}, which automatically generates rankings by fitting a \gls{bc} policy on the demonstrations and subsequently draws samples with different levels of noise from this policy.
We adapt \gls{drex} to our trajectory-based setting as a baseline.
Similar to \gls{vigor}, this adapted method trains a model on expert demonstrations on training contexts to optimize \gls{gmm} policies on novel test contexts. 
However, while \gls{drex} learns a reward function to do so, \gls{vigor} iteratively (re-)trains a discriminator on the expert demonstrations and policy samples.

\section{Foundations}
In this section, we briefly cover the foundations and previous work that our method builds on.

\textbf{Probabilistic Movement Primitives.}
For a trajectory $\btau=(\btau_1, \cdots, \btau_T)$, 
\glspl{promp} represent the elements as
$\btau_t = \bm{\Phi}(t)^T\w$, where $\btau_t$ represents a vector of the desired joint angles at time-step $t$~\citep{zhou2020movement} Note that $\w$ does not depends on $t$ and that $\btau$ can thus be computed at an arbitrary resolution $T$.
Here $\bm{\Phi}(t)$ are time-dependent features, usually radial basis functions centered around different time points, and $\w$ denotes the parameter vector. 
For a single demonstration, the parameters $\w$ are fitted using simple linear regression and compactly represent the entire trajectory.

\textbf{Variational Inference for Gaussian Mixture Models.}
We build on a recent class of efficient Variational Inference approaches that allow modeling versatile behavior with \glspl{gmm}~\citep{Becker2020Expected,arenz2018efficient}.
These approaches use the Information-Projection~\citep{murphy2012machine} to minimize the reverse Kullback-Leibler Divergence $\textrm{KL}\big(q(\w) || p (\w)\big)$ between a model $q(\w)$ and a target distribution $p(\w)$. 
In our case, $p(\w)$ is the distribution of expert trajectories.
\citet{arenz2018efficient} introduced an upper-bound objective based on a variational decomposition.
The resulting approach, \gls{vips}~\citep{arenz2018efficient, arenz2020trust}, repeatedly solves the objective 
\begin{align}
\label{eq:vips_objective}
 q\left( \w\right) =\argmin_{q\left(\w\right)}
\mathbb{E}_{q\left(\w, \z\right)}
\left[ \log \dfrac{\hat{q}\left(\w\right)}{p\left(\w\right)}\right] + \text{KL}\big(q(\z) || \hat{q}(\z)\big) + \mathbb{E}_{q\left(\z\right)}\Big[\text{KL}\big(q\left(\w|\z\right) || \hat{q}\left(\w|\z\right)\big)\Big],
\end{align}
where $\hat{q}$ denotes the model from the previous iteration. 
This objective decomposes into individual optimizations for the components and categorical distributions, which~\citet{arenz2018efficient} solve using trust-region methods from policy search~\citep{peters2010relative, abdolmaleki2015model}.

\textbf{Distribution Matching for Gaussian Mixture Models.}
We cannot directly work with~\gls{vips}, as we do not know $p(\w)$ but only have the expert demonstrations. 
To address this issue,~\citet{Becker2020Expected} propose using density ratio estimation~\citep{sugiyama2012density} approaches to approximate the $\log (\hat{q}(\w) / p(\w)) $ term in Equation \ref{eq:vips_objective}, removing the dependency on $p(\w)$.
Specifically, they use logistic regression to estimate the density ratio.
The resulting approach, \gls{eim}, resembles Generative Adversarial Nets~\citep{goodfellow2014generative} where the discriminator effectively also estimates a density ratio.


\section{Inferring Versatile Behaviors by Matching Geometric Descriptors}
\label{sec:method}
Our approach estimates versatile behavior in form of \gls{gmm} distributions over \gls{promp} weights for all training and test contexts of the given task.
We note that while this setup requires the test contexts in advance, 
it does not need interaction with the real environment since the behavioral descriptors are directly computed from the proposed trajectory and its context.
To capture correlations in the demonstrations, we utilize full-covariance \gls{gmm} components.
For each planned trajectory, we compute concise descriptors such as distances to key-points that isolate the performance of the trajectory from its context to allow for sample-efficient generalization to novel contexts.
We then build on \gls{eim}~\citep{Becker2020Expected} and use a discriminator to infer \glspl{gmm} from demonstrations by matching the learner's and expert's distributions in the resulting geometric feature space. Towards this end, we generalize \gls{eim} to sequential data and modify the policy parameterization to accommodate for our setting.
Figure \ref{fig:schematic} illustrates an overview of our approach.
We provide pseudocode in Appendix \ref{app_sec:pseudocode}.

\begin{figure}[t]
\centering
    \includegraphics[width=\textwidth]{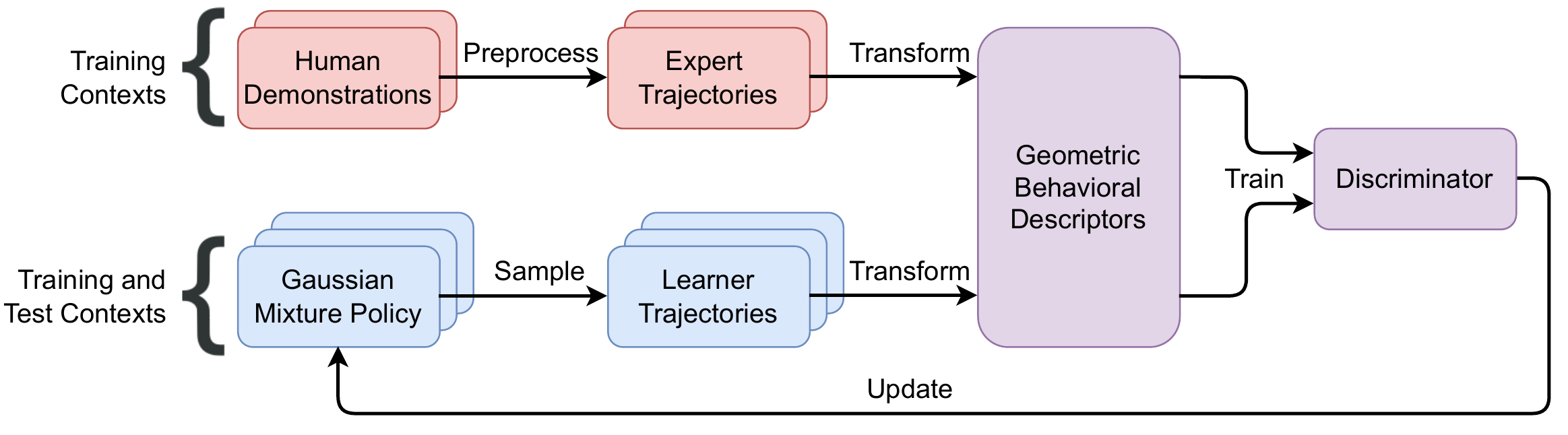}
    \caption{Schematic of \gls{vigor}.  
    We generate a set of expert trajectories from human demonstrations and then transform them into geometric behavioral descriptors. These descriptors are then fed into a discriminator. 
    This process is repeated for samples of a separate \gls{gmm} policy for each training and test context.
    We then train the discriminator to distinguish between human and policy trajectories.
    The geometric descriptors are designed to abstract away from concrete task configurations, causing the discriminator to distinguish how well a trajectory performs rather than which context it acts on. As a result, the discriminator can be used to improve the policies.
    }
    \label{fig:schematic}
\end{figure}

\textbf{Context-Specific Mixture Policies}
We want to model the expert's behavior for a given small set of training configurations of the task, denoted as training contexts $\ctx_{\textrm{train}}$, and a set of unseen test contexts $\ctx_{\textrm{test}}$. We consider the set of all contexts as $\ctx = \ctx_{\textrm{train}}\cup \ctx_{\textrm{test}}$.
As context-dependent \glspl{gmm}\footnote{Such \glspl{gmm} could in principle be represented by \glspl{mdn}. However, attempts with \gls{mdn} and full-covariance Gaussian components were highly unsuccessful in our experiments.} are hard to learn and we assume a relatively small amount of training contexts, $6$ in all our experiments, with multiple demonstrations per training context, we resort to maintaining a single \gls{gmm}-Policy $q_{c}(\w)$ for each context $c$. 
This non-amortized formulation simplifies the policy updates as we can optimize the policy for each context individually, given the density ratio estimator, using highly efficient, tailored methods for full-rank \gls{gmm} approximations~\citep{arenz2018efficient,arenz2020trust}. 

\textbf{Distribution Matching of Behavioral Descriptors}
We assume access to behavioral descriptors $\Obs = f_c(\w)$ that represent features encoded in the parameter vector $\w$ with respect to the context $c$ of $\w$. 
For simplicity, we will assume that the mapping $f_c$ between $\w$ and $\Obs$ is deterministic, such that we can easily evaluate $\Obs$ for any desired plan $\w$.
We want to match the distribution of behavioral descriptors of the demonstrator while optimizing for $q(\bm{w})$, i.e,
\begin{align*}
    q^*(\w) = \argmin_{q(\w)} \textrm{KL}\big(q\left(\Obs\right)||p(\Obs)\big) = \argmin_{q(\w)} \E{q(\w)}{\int_{\Obs} p(\bm{O}|\bm{w}) \log \frac{q(\bm{O})}{p(\bm{O})} d\bm{O}},
\end{align*}
 where $p(\Obs|\w)$ is a Dirac delta distribution defined by the mapping $f_c$ and $q(\bm{O}) = \int p(\Obs|\w) q(\bm{w})d \w $. As the mapping between $\w$ and $\Obs$ is deterministic, the \gls{eim} algorithm can be directly applied to this setup with the difference that the discriminator is trained using $\Obs$ instead of $\w$. 
 Moreover, while we a learn different distributions $q_{c}(\w)$ for each context $c$, we use the same discriminator for all contexts.
This way, we can infer distributions $\{q_c(\w) | c\in {\ctx_{\textrm{test}}}\}$, i.e. infer versatile trajectories for unseen scenarios.
 
\textbf{Density Ratio Estimation for Sequential Behavioral Descriptors}
As in our case, $\w$ encodes the desired trajectory $\btau$, the behavioral descriptors are typically computed per time-step, i.e., $\Obs = (\obs_1, \dots, \obs_T)^T$, where each $\obs_t$ is a feature vector.
To enable the classifier to deal with sequential data, we consider a sequence-to-sequence neural network $\bphi(\Obs)=\bphi((\obs_1, \dots, \obs_T)^T) = (y_1, \dots, y_T)^T$. The network receives a sequence of inputs $\Obs=(\obs_1,\dots, \obs_T)^T$ and outputs a sequence of values $(y_1, \dots, y_T)^T$, where each $y_i$ may depend on multiple $\obs_j$.
In each iteration of our optimization, we (re-)train this network to discriminate between the provided demonstrations $\Obs^{(p)}$ of $\ctx_{\textrm{train}}$, and an equal number of policy samples $\Obs^{(q)}$ drawn uniformly from $\ctx$. We first consider the case where we want to discriminate full trajectories as in \gls{eim}.
Denoting the sigmoid function as $\sigma$, the discriminator can straightforwardly be trained on a binary cross-entropy loss of the sum of sequence values $\hat{y}= \sum_{t=0}^T y_t$, i.e., by minimizing
\begin{align}
\label{eq:full_bce}
    \text{BCE}\big(\bphi(\Obs), \Obs^{(p)}, \Obs^{(q)}\big) 
    & = -\E{\Obs^{(q)}}{\log\big(\sigma\left(\hat{y}\right)\big)} 
    -\E{\Obs^{(p)}}{\log\big(1-\sigma\left(\hat{y}\right)\big)},
\end{align}
w.r.t. $\bphi(\Obs)$. Equation \ref{eq:full_bce} recovers the log density ratio $\bphi(\Obs)=\log \Obs^{(p)} - \log \Obs^{(q)}$ at convergence~\citep{sugiyama2012density}. However, this cost function only provides a single training sample per trajectory and is therefore hard to use for a small number of demonstrations.  
By utilizing the sequential structure of the trajectories, the classification error can be computed for each time-step, i.e. we minimize
\begin{align}
\label{eq:stepwise_bce}
    \text{BCE}_{\text{step}}\big(\bphi(\Obs), \Obs^{(p)}, \Obs^{(q)}\big) 
    & = -\E{\Obs^{(q)}}{\sum_{t=0}^T\log\big(\sigma\left( y_t\right)\big)} 
    -\E{\Obs^{(p)}}{\sum_{t=0}^T \log\big(1-\sigma\left(y_t\right)\big)}
\end{align}
w.r.t. $\bphi(\Obs)$.
Finally, we train an ensemble of discriminators instead of a single one, using their average logit as the log density ratio estimate for the policy updates.
For the network architecture, we find that simple $2-4$ layer $1d$ convolutional neural networks ($1d$-CNNs) work best in our experiments. We compare this choice to other sequential network architectures the ablations in Appendix \ref{app_sec:ablations}.

\textbf{Geometric Behavioral Descriptors}
\label{ssec:geometric_descriptors}
\gls{vigor} utilizes geometric descriptors that abstract away from a concrete task configuration to generalize to new configurations. 
To this end, we encode the current state along the desired trajectory with respect to relative geometric features rather than absolute values. 
The resulting geometric descriptors cause trajectories of similar performance to appear similar in feature space regardless of their context, allowing for sample-efficient generalization to novel contexts.
We note that such a descriptor space can often be straightforwardly constructed from the geometry of the task by composing distances of the end-effector to key-points of the objects in a scene.
For example, in object manipulation, this space can be composed of distances of the end-effector to the corners of a box to push.

\section{Experiments}
\label{sec:results}
All experiments use human demonstrations. The tasks were selected due to their highly multi-modal nature and because they allow for an easy collection of human demonstrations. We instructed the demonstrators to solve the task in varying ways to create versatile solutions.
As a preprocessing step, we fit a \gls{promp} for each human demonstration and filter the resulting \glspl{promp} such that only successful demonstrations remain.
This step ensures that the human demonstrations can be imitated by the learner and conveniently allows to compress all demonstrations to a fixed length in a principled fashion.
We use the resulting preprocessed expert demonstrations for all experiments unless otherwise stated.
Similarly, unless noted otherwise, all experiments use $6$ training and $6$ test contexts, and $5$ \gls{gmm} components for each configuration and method.
Experimentally, this number of components is sufficient for matching the distribution of expert descriptors for the considered tasks, whereas more components generally only lead to spurious improvements. 
We experiment with less components in the ablations in Appendix \ref{app_sec:ablations}. Unless noted otherwise, we do \textit{not} train the categorical distribution of the \gls{gmm} components, using a uniform distribution instead.
We evaluate the \gls{gmm} policies using a number of samples for each component of each \textit{test} context, and then rank these components according to the average performance of their samples. 
We then report statistics of the best component of the resulting value over the random seeds.
Note that reporting the best component leads to a fair comparison to unimodal approaches.
For baselines that do not use \gls{gmm} policies, we instead draw samples of the trained policies for each test context and average the performance of all of them. 
For each experiment, we report a measure of how well the task is performed in the main paper, and additional success rates in Appendix \ref{app_sec:task_info_and_results}.
We repeat each experiment for $10$ random seeds.
Appendix \ref{app_sec:hyperparameters} shows all hyperparameters and training details.

\textbf{Baselines.}
We compare \gls{vigor} to a number of \gls{il} baselines, using three state-action baselines, namely \gls{gail}~\citep{ho2016generative}, behavioral cloning (BC(S)) and behavioral cloning with a \gls{gmm} policy (BC-GMM(S))~\citep{mandlekar2021matters}. For all experiments, the state-action baselines receive geometric descriptors of the current state and a time-step as state information, and output the velocities of the robot joints as actions.
Further, we also consider three trajectory-based methods. These are standard behavioral cloning (BC(T)), behavioral cloning with a \gls{gmm} policy (BC-GMM(T)) (see e.g., \citet{zhou2020movement}), and a modified version of D-REX (EM+D-REX). 
The latter is chosen due to its similarities to \gls{vigor}, which are detailed in Appendix \ref{app_ssec:em_drex}.
We refer to Appendix \ref{app_sec:baselines} for details on the baselines and to Appendix \ref{app_sec:task_info_and_results} for further results on all experiments described below.

\begin{figure}[t]
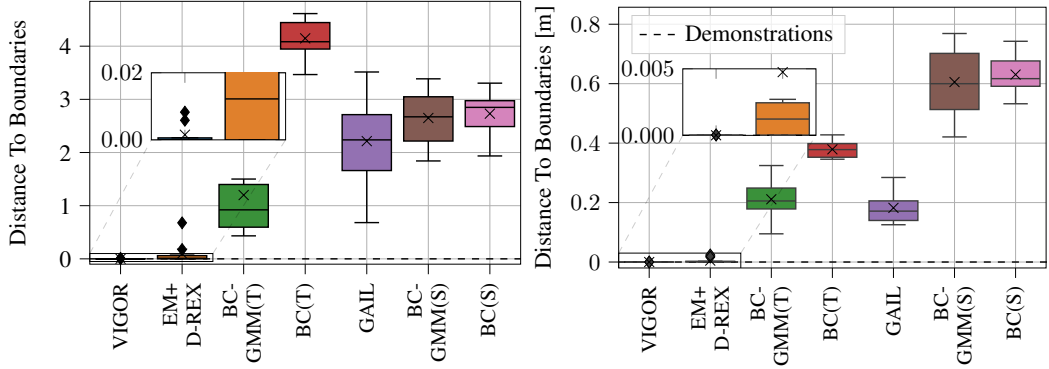

    \centering
    \begin{minipage}{0.5\textwidth}
        \resizebox{\textwidth}{!}{\input{figures/results/mpr/distance_to_boundaries_test}}
    \end{minipage}%
    \begin{minipage}{0.5\textwidth}
        \resizebox{\textwidth}{!}{\input{figures/results/tel/distance_to_boundaries_test}}
    \end{minipage}
    \caption{Mean target distance of samples from the best policy component on the Planar (left) and Panda (right) Reaching tasks for test contexts. 
    We find that a trajectory-based setting and the use of mixture policies help performance, as can be seen from \gls{vigor}, EM+D-REX, and BC-GMM (T). Similarly, using a discriminator to iteratively optimize the policies is advantageous on the reaching tasks, as seen from \gls{vigor} and GAIL. Finally, the clear distribution matching objective of \gls{vigor} and EM+D-REX seems crucial for imitation from a low number of versatile demonstrations.
    \gls{vigor} uniquely combines the above properties and is able to reliably reach both targets for both tasks.
    }
    \label{fig:results_reacher_tasks}
\end{figure}

\textbf{Planar Reacher.}
First, we consider the introductory task shown in Figure \ref{fig:figure_one}.
In this task, the goal is to reach an intermediate target area with the end-effector of a $5$ \gls{dof} robot with joints of length $1$ before ending the trajectory in a final target area.
The geometric descriptors for this task are given by the euclidean distance of the end-effector to both targets, paired with the average velocity and acceleration of the joints to encode the smoothness of the motion. This yields a total of $4$ features per time-step. 
The left and middle panels of Figure \ref{fig:figure_one} depict example demonstrations for two training contexts (top) and the corresponding distances between the end-effector and targets over time (bottom).
We explore other choices for the geometric descriptors in the ablations in Appendix \ref{app_sec:ablations}.
For evaluation, we use the \textit{Distance to Boundaries} of the target areas. For the first target, we consider the minimum distance of the end-effector to the boundary; for the second target, we consider the distance at the last time-step.
We train all methods on $5$ demonstrations per training context.
The left of Figure \ref{fig:results_reacher_tasks} shows results on test contexts. 
We find that only \gls{vigor} and \gls{em}+\gls{drex} can consistently get close to both targets. Yet, \gls{em}+\gls{drex} can only learn a single mode and fails to reproduce the versatility of the demonstrations.
We investigate this in more detail in Appendix \ref{app_sec:planar_point_reacher}.

\textbf{$7$-D Panda Reacher.}
\label{sec:panda_reacher}
We repeat the above point-reaching task using a $7$ \gls{dof} Franka Emika Panda robot and $3$-D intermediate and goal positions.
Since we do not care about the end-effector's rotation, the last \gls{dof} can be ignored, leading to an effective action dimension of $6$.
We use $8$ basis functions per action dimension for the \glspl{mp}, leading to a total of \gls{mp} dimension of $48$ and thus a high-dimensional problem space.
The expert demonstrations are collected using a teleoperation setup with a virtual twin that records joint values over time. 
Appendix \ref{app_sec:panda_reacher} explains the setup in more detail.
We use the same geometric descriptors as for the planar reacher task.
The right of Figure \ref{fig:results_reacher_tasks} shows target distances on test contexts.
Similar to the previous task, only \gls{vigor} and \gls{em}+\gls{drex} can consistently get close to both targets while the other methods regularly fail to pursue at least one of the two targets.

\textbf{Box Pusher.}
\label{sec:box_pusher}
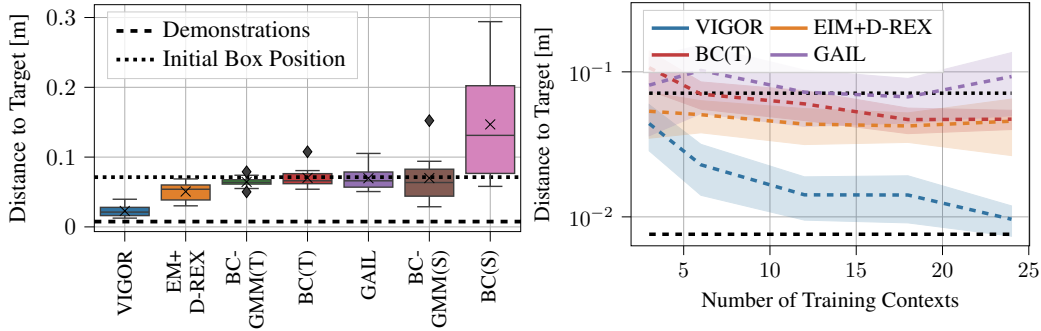
\begin{figure}[t]
    \begin{minipage}[t]{0.5\textwidth}
    \vspace{0pt}
    \resizebox{\textwidth}{!}{
\begin{tikzpicture}

\definecolor{brown1926061}{RGB}{192,60,61}
\definecolor{darkgray176}{RGB}{176,176,176}
\definecolor{darkslategray61}{RGB}{61,61,61}
\definecolor{dimgray1319183}{RGB}{131,91,83}
\definecolor{forestgreen4416044}{RGB}{0,0,0} 
\definecolor{lightgray204}{RGB}{204,204,204}
\definecolor{mediumpurple147113178}{RGB}{147,113,178}
\definecolor{orchid213132188}{RGB}{213,132,188}
\definecolor{peru22412844}{RGB}{224,128,44}
\definecolor{seagreen5814558}{RGB}{58,145,58}
\definecolor{steelblue49115161}{RGB}{49,115,161}

\begin{axis}[
legend cell align={left},
legend style={
  fill opacity=0.8,
  draw opacity=1,
  text opacity=1,
  at={(0.03,0.97)},
  anchor=north west,
  draw=lightgray204
},
width=8cm,
height=5.0cm,
tick align=outside,
tick pos=left,
x grid style={darkgray176},
xmajorgrids,
xmin=-0.5, xmax=6.5,
xtick style={color=black},
xtick={0,1,2,3,4,5,6},
xticklabels={VIGOR,EM+\\D-REX,BC-\\GMM(T),BC(T),GAIL,BC-\\GMM(S),BC(S)},
y grid style={darkgray176},
ylabel={Distance to Target [m]},
ymajorgrids,
ymin=-0.00673630649997203, ymax=0.318404428499413,
ytick style={color=black},
x tick label style={rotate=90, font=\small, align=right}
]
\path [draw=darkslategray61, fill=steelblue49115161, semithick]
(axis cs:-0.4,0.0161759620731229)
--(axis cs:0.4,0.0161759620731229)
--(axis cs:0.4,0.0279913769060086)
--(axis cs:-0.4,0.0279913769060086)
--(axis cs:-0.4,0.0161759620731229)
--cycle;
\path [draw=darkslategray61, fill=peru22412844, semithick]
(axis cs:0.6,0.0385798133601263)
--(axis cs:1.4,0.0385798133601263)
--(axis cs:1.4,0.0600547309852459)
--(axis cs:0.6,0.0600547309852459)
--(axis cs:0.6,0.0385798133601263)
--cycle;
\path [draw=darkslategray61, fill=seagreen5814558, semithick]
(axis cs:1.6,0.0613152464518539)
--(axis cs:2.4,0.0613152464518539)
--(axis cs:2.4,0.0675097177106635)
--(axis cs:1.6,0.0675097177106635)
--(axis cs:1.6,0.0613152464518539)
--cycle;
\path [draw=darkslategray61, fill=brown1926061, semithick]
(axis cs:2.6,0.0619183815339838)
--(axis cs:3.4,0.0619183815339838)
--(axis cs:3.4,0.0762975388645375)
--(axis cs:2.6,0.0762975388645375)
--(axis cs:2.6,0.0619183815339838)
--cycle;
\path [draw=darkslategray61, fill=mediumpurple147113178, semithick]
(axis cs:3.6,0.0569290115426822)
--(axis cs:4.4,0.0569290115426822)
--(axis cs:4.4,0.0786498440254862)
--(axis cs:3.6,0.0786498440254862)
--(axis cs:3.6,0.0569290115426822)
--cycle;
\path [draw=darkslategray61, fill=dimgray1319183, semithick]
(axis cs:4.6,0.0439994674769161)
--(axis cs:5.4,0.0439994674769161)
--(axis cs:5.4,0.0826372911012431)
--(axis cs:4.6,0.0826372911012431)
--(axis cs:4.6,0.0439994674769161)
--cycle;
\path [draw=darkslategray61, fill=orchid213132188, semithick]
(axis cs:5.6,0.0766449339494102)
--(axis cs:6.4,0.0766449339494102)
--(axis cs:6.4,0.202271902514666)
--(axis cs:5.6,0.202271902514666)
--(axis cs:5.6,0.0766449339494102)
--cycle;
\addplot [ultra thick, black, dashed]
table {%
-0.5 0.00758827236363636
7.5 0.00758827236363636
};
\addlegendentry{Demonstrations}
\addplot [ultra thick, black, dotted]
table {%
-0.5 0.0712566969090909
7.5 0.0712566969090909
};
\addlegendentry{Initial Box Position}
\addplot [semithick, darkslategray61, forget plot]
table {%
0 0.0161759620731229
0 0.0125687005796863
};
\addplot [semithick, darkslategray61, forget plot]
table {%
0 0.0279913769060086
0 0.0396162831927243
};
\addplot [semithick, darkslategray61, forget plot]
table {%
-0.2 0.0125687005796863
0.2 0.0125687005796863
};
\addplot [semithick, darkslategray61, forget plot]
table {%
-0.2 0.0396162831927243
0.2 0.0396162831927243
};
\addplot [semithick, darkslategray61, forget plot]
table {%
1 0.0385798133601263
1 0.030181163020761
};
\addplot [semithick, darkslategray61, forget plot]
table {%
1 0.0600547309852459
1 0.0687314488871354
};
\addplot [semithick, darkslategray61, forget plot]
table {%
0.8 0.030181163020761
1.2 0.030181163020761
};
\addplot [semithick, darkslategray61, forget plot]
table {%
0.8 0.0687314488871354
1.2 0.0687314488871354
};
\addplot [semithick, darkslategray61, forget plot]
table {%
2 0.0613152464518539
2 0.054987727603205
};
\addplot [semithick, darkslategray61, forget plot]
table {%
2 0.0675097177106635
2 0.0739834701177484
};
\addplot [semithick, darkslategray61, forget plot]
table {%
1.8 0.054987727603205
2.2 0.054987727603205
};
\addplot [semithick, darkslategray61, forget plot]
table {%
1.8 0.0739834701177484
2.2 0.0739834701177484
};
\addplot [black, mark=diamond*, mark size=2.5, mark options={solid,fill=darkslategray61}, only marks, forget plot]
table {%
2 0.0501747953608339
2 0.0789727479021223
};
\addplot [semithick, darkslategray61, forget plot]
table {%
3 0.0619183815339838
3 0.0539727796089806
};
\addplot [semithick, darkslategray61, forget plot]
table {%
3 0.0762975388645375
3 0.0806509072492046
};
\addplot [semithick, darkslategray61, forget plot]
table {%
2.8 0.0539727796089806
3.2 0.0539727796089806
};
\addplot [semithick, darkslategray61, forget plot]
table {%
2.8 0.0806509072492046
3.2 0.0806509072492046
};
\addplot [black, mark=diamond*, mark size=2.5, mark options={solid,fill=darkslategray61}, only marks, forget plot]
table {%
3 0.107660750157418
};
\addplot [semithick, darkslategray61, forget plot]
table {%
4 0.0569290115426822
4 0.050624376560793
};
\addplot [semithick, darkslategray61, forget plot]
table {%
4 0.0786498440254862
4 0.105241655577567
};
\addplot [semithick, darkslategray61, forget plot]
table {%
3.8 0.050624376560793
4.2 0.050624376560793
};
\addplot [semithick, darkslategray61, forget plot]
table {%
3.8 0.105241655577567
4.2 0.105241655577567
};
\addplot [semithick, darkslategray61, forget plot]
table {%
5 0.0439994674769161
5 0.0288556143145606
};
\addplot [semithick, darkslategray61, forget plot]
table {%
5 0.0826372911012431
5 0.0940035157904506
};
\addplot [semithick, darkslategray61, forget plot]
table {%
4.8 0.0288556143145606
5.2 0.0288556143145606
};
\addplot [semithick, darkslategray61, forget plot]
table {%
4.8 0.0940035157904506
5.2 0.0940035157904506
};
\addplot [black, mark=diamond*, mark size=2.5, mark options={solid,fill=darkslategray61}, only marks, forget plot]
table {%
5 0.152269407629126
};
\addplot [semithick, darkslategray61, forget plot]
table {%
6 0.0766449339494102
6 0.058046877257729
};
\addplot [semithick, darkslategray61, forget plot]
table {%
6 0.202271902514666
6 0.294079849635804
};
\addplot [semithick, darkslategray61, forget plot]
table {%
5.8 0.058046877257729
6.2 0.058046877257729
};
\addplot [semithick, darkslategray61, forget plot]
table {%
5.8 0.294079849635804
6.2 0.294079849635804
};
\addplot [semithick, darkslategray61, forget plot]
table {%
-0.4 0.0212062108013391
0.4 0.0212062108013391
};
\addplot [forestgreen4416044, mark=x, mark size=3, mark options={solid,fill=black}, only marks, forget plot]
table {%
0 0.0228599832489933
};
\addplot [semithick, darkslategray61, forget plot]
table {%
0.6 0.0539347518881532
1.4 0.0539347518881532
};
\addplot [forestgreen4416044, mark=x, mark size=3, mark options={solid,fill=black}, only marks, forget plot]
table {%
1 0.0506390572288727
};
\addplot [semithick, darkslategray61, forget plot]
table {%
1.6 0.0639072114718547
2.4 0.0639072114718547
};
\addplot [forestgreen4416044, mark=x, mark size=3, mark options={solid,fill=black}, only marks, forget plot]
table {%
2 0.0644310424949681
};
\addplot [semithick, darkslategray61, forget plot]
table {%
2.6 0.0660944511816781
3.4 0.0660944511816781
};
\addplot [forestgreen4416044, mark=x, mark size=3, mark options={solid,fill=black}, only marks, forget plot]
table {%
3 0.0702370889008204
};
\addplot [semithick, darkslategray61, forget plot]
table {%
3.6 0.0660149520445226
4.4 0.0660149520445226
};
\addplot [forestgreen4416044, mark=x, mark size=3, mark options={solid,fill=black}, only marks, forget plot]
table {%
4 0.0697225520008415
};
\addplot [semithick, darkslategray61, forget plot]
table {%
4.6 0.0635561578941002
5.4 0.0635561578941002
};
\addplot [forestgreen4416044, mark=x, mark size=3, mark options={solid,fill=black}, only marks, forget plot]
table {%
5 0.069929637400071
};
\addplot [semithick, darkslategray61, forget plot]
table {%
5.6 0.131145074915677
6.4 0.131145074915677
};
\addplot [forestgreen4416044, mark=x, mark size=3, mark options={solid,fill=black}, only marks, forget plot]
table {%
6 0.146857389081734
};
\end{axis}

\end{tikzpicture}}
    \end{minipage}%
    \begin{minipage}[t]{0.5\textwidth}
    \vspace{0pt}
    \resizebox{\textwidth}{!}{
\begin{tikzpicture}

\definecolor{crimson2143940}{RGB}{214,39,40}
\definecolor{darkgray176}{RGB}{176,176,176}
\definecolor{darkorange25512714}{RGB}{255,127,14}
\definecolor{lightgray204}{RGB}{204,204,204}
\definecolor{mediumpurple147113178}{RGB}{147,113,178}
\definecolor{steelblue49115161}{RGB}{49,115,161}
\definecolor{orchid213132188}{RGB}{213,132,188}
\definecolor{peru22412844}{RGB}{224,128,44}
\definecolor{brown1926061}{RGB}{192,60,61}
\definecolor{darkgray176}{RGB}{176,176,176}
\definecolor{darkslategray61}{RGB}{0,0 0} 
\definecolor{dimgray1319183}{RGB}{131,91,83}
\definecolor{forestgreen4416044}{RGB}{0,0,0}
\definecolor{lightgray204}{RGB}{204,204,204}
\definecolor{mediumpurple147113178}{RGB}{147,113,178}
\definecolor{seagreen5814558}{RGB}{58,145,58}
\definecolor{steelblue49115161}{RGB}{49,115,161}

\begin{axis}[
legend cell align={left},
legend style={
  fill opacity=0.8,
  draw opacity=1,
  text opacity=1,
  at={(0.03,0.97)},
  anchor=north west,
  legend columns=2,
  draw=lightgray204
},
log basis y={10},
width=8cm,
height=5.5cm,
tick align=outside,
tick pos=left,
x grid style={darkgray176},
xlabel={Number of Training Contexts},
xmajorgrids,
xmin=1.95, xmax=25.05,
xtick style={color=black},
y grid style={darkgray176},
ylabel={Distance to Target [m]},
ymajorgrids,
ymin=0.00631610358542816, ymax=0.3, 
ymode=log,
ytick style={color=black},
ytick={0.0001,0.001,0.01,0.1,1,10},
yticklabels={
  \(\displaystyle {10^{-4}}\),
  \(\displaystyle {10^{-3}}\),
  \(\displaystyle {10^{-2}}\),
  \(\displaystyle {10^{-1}}\),
  \(\displaystyle {10^{0}}\),
  \(\displaystyle {10^{1}}\)
}
]
\addlegendimage{steelblue49115161, ultra thick}
\addlegendentry{VIGOR}
\addlegendimage{peru22412844, ultra thick}
\addlegendentry{EIM+D-REX}
\addlegendimage{brown1926061, ultra thick}
\addlegendentry{BC(T)}
\addlegendimage{mediumpurple147113178, ultra thick}
\addlegendentry{GAIL}

\path [draw=steelblue49115161, fill=steelblue49115161, opacity=0.2]
(axis cs:3,0.0594549307242403)
--(axis cs:3,0.0285233073218294)
--(axis cs:6,0.0140811652951976)
--(axis cs:12,0.0094934526330449)
--(axis cs:18,0.00908664854982778)
--(axis cs:24,0.00735160090980173)
--(axis cs:24,0.011904664105725)
--(axis cs:24,0.011904664105725)
--(axis cs:18,0.019207620748508)
--(axis cs:12,0.0188847710409027)
--(axis cs:6,0.031638801202789)
--(axis cs:3,0.0594549307242403)
--cycle;

\path [draw=peru22412844, fill=peru22412844, opacity=0.2]
(axis cs:3,0.072116376109381)
--(axis cs:3,0.034864177175427)
--(axis cs:6,0.0379932206000978)
--(axis cs:12,0.0315231443003849)
--(axis cs:18,0.0326141418440199)
--(axis cs:24,0.0264597038085138)
--(axis cs:24,0.0649223180340467)
--(axis cs:24,0.0649223180340467)
--(axis cs:18,0.0521098409928075)
--(axis cs:12,0.0558016934953578)
--(axis cs:6,0.0632848938576475)
--(axis cs:3,0.072116376109381)
--cycle;

\path [draw=brown1926061, fill=brown1926061, opacity=0.2]
(axis cs:3,0.139263234705485)
--(axis cs:3,0.074405801020528)
--(axis cs:6,0.0552140598732987)
--(axis cs:12,0.0461520715486417)
--(axis cs:18,0.0374837914730677)
--(axis cs:24,0.0400137351040946)
--(axis cs:24,0.0543186753598163)
--(axis cs:24,0.0543186753598163)
--(axis cs:18,0.056248113130066)
--(axis cs:12,0.0741357573308033)
--(axis cs:6,0.0852601179283421)
--(axis cs:3,0.139263234705485)
--cycle;

\path [draw=mediumpurple147113178, fill=mediumpurple147113178, opacity=0.2]
(axis cs:3,0.125586517513127)
--(axis cs:3,0.0362412476299226)
--(axis cs:6,0.0515750180582571)
--(axis cs:12,0.0420471068463124)
--(axis cs:18,0.0443783358653302)
--(axis cs:24,0.0498665689065641)
--(axis cs:24,0.135791490955037)
--(axis cs:24,0.135791490955037)
--(axis cs:18,0.0900666946597829)
--(axis cs:12,0.102782269966993)
--(axis cs:6,0.153121262508505)
--(axis cs:3,0.125586517513127)
--cycle;

\addplot [ultra thick, black, dashed, forget plot]
table {%
3 0.00758827236363636
24 0.00758827236363636
};
\addplot [ultra thick, black, dotted, forget plot]
table {%
3 0.0712566969090909
24 0.0712566969090909
};
\addplot [ultra thick, steelblue49115161, dashed, forget plot]
table {%
3 0.0439891190230348
6 0.0228599832489933
12 0.0141891118369738
18 0.0141471346491679
24 0.00962813250776337
};
\addplot [ultra thick, peru22412844, dashed, forget plot]
table {%
3 0.053490276642404
6 0.0506390572288727
12 0.0436624188978714
18 0.0423619914184137
24 0.0456910109212803
};
\addplot [ultra thick, brown1926061, dashed, forget plot]
table {%
3 0.106834517863007
6 0.0702370889008204
12 0.0601439144397225
18 0.0468659523015669
24 0.0471662052319555
};
\addplot [ultra thick, mediumpurple147113178, dashed, forget plot]
table {%
3 0.0809138825715248
6 0.102348140283381
12 0.0724146884066527
18 0.0672225152625566
24 0.0928290299308004
};
\end{axis}

\end{tikzpicture}}
    \end{minipage}
    \caption{Mean target distance of samples from the best policy component on the Box Pushing task for test contexts.
    The dotted line denotes the average corner distance between the initial and the desired box position. 
    The dashed line shows the performance of the human demonstrations. 
    (Left) \gls{vigor} is able to consistently push the box to the right configuration.
    Opposed to this, the other methods only marginally improve over the initial box positions, and in some cases perform worse than it.
    (Right) We find that \gls{vigor} benefits most from additional training contexts, almost matching the performance of the demonstrator on unseen contexts for $24$ training contexts.}
    \label{fig:results_box_pusher}
\end{figure}
We also evaluate our approach on a contact-rich robot manipulation task. 
In this task, a simulated Franka Emika Panda robot has a rod attached to its end-effector and needs to use it to push a rectangular box to a given goal position and orientation.
The box always starts at the same position, and the task context is given by the desired $(x,y)$ translation of the box, plus its desired rotation in degrees. 
For simplicity, the task is learned in task space with a fixed height. 
The demonstrations contain the starting position of the robot as well as its trajectory. Here, the solutions are versatile because the box can be pushed both from the inside and outside.
The geometric descriptors consist of the $(x,y)$ distances of the end-effector to the \textit{initial} and \textit{desired final} position of the corners of the box. 
To facilitate generalization between contexts, we translate the basis of these distances such that the desired final box aligns with the origin of the coordinate system.
We additionally add the end-effector velocity and acceleration as well as the time-step, resulting in $19$ features per time-step.
This geometric description of the task allows for \gls{il} without simulating or explicitly modeling the box, significantly speeding up the training process and facilitating generalization to both new contexts and real-world robots. 
We use $3$ demonstrations per training context and evaluate the average distance of the corners of the final box to that of the \textit{desired final} position.
The contact with the box causes a mismatch between a planned trajectory and its execution on the robot, which we illustrate in Appendix \ref{app_sec:box_pusher}.
Results are shown in the left of Figure \ref{fig:results_box_pusher}. Additionally, the right of Figure \ref{fig:results_box_pusher} shows how the performance varies when changing the number of training contexts.
Finally, we illustrate how the planned trajectories of \gls{vigor} perform on a real robot. To this end, we create a real-world replica of the simulated box that the robot manipulates.
Rollouts for different policy components of \gls{vigor} trained on $24$ training contexts on an exemplary test context are shown in Figure~\ref{fig:box_pusher_real_robots}.

\begin{figure}[t]	
  \begin{minipage}[t!]{\textwidth}
  \begin{minipage}[t!]{\textwidth}
	\begin{minipage}[t!]{0.138\textwidth}
		\includegraphics[width=\textwidth]{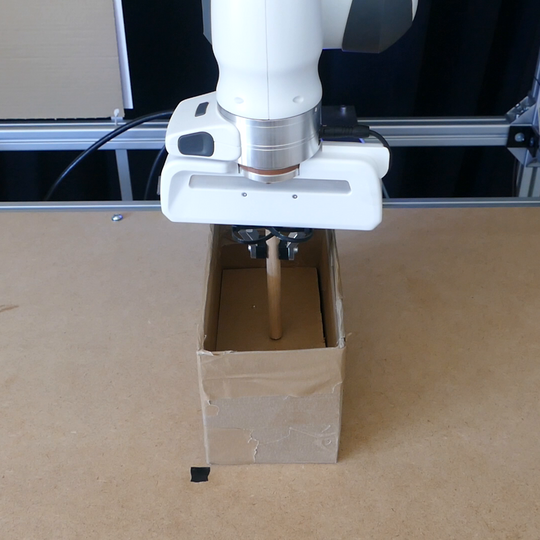}
	\end{minipage}\hfill
	\begin{minipage}[t!]{0.138\textwidth}
		\includegraphics[width=\textwidth]{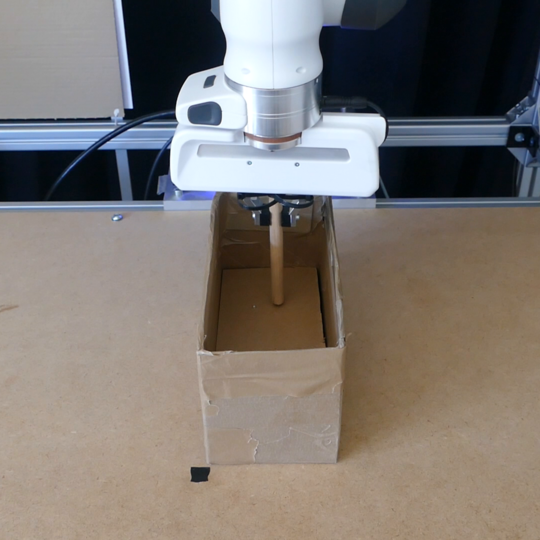}
	\end{minipage}\hfill
	\begin{minipage}[t!]{0.138\textwidth}
		\includegraphics[width=\textwidth]{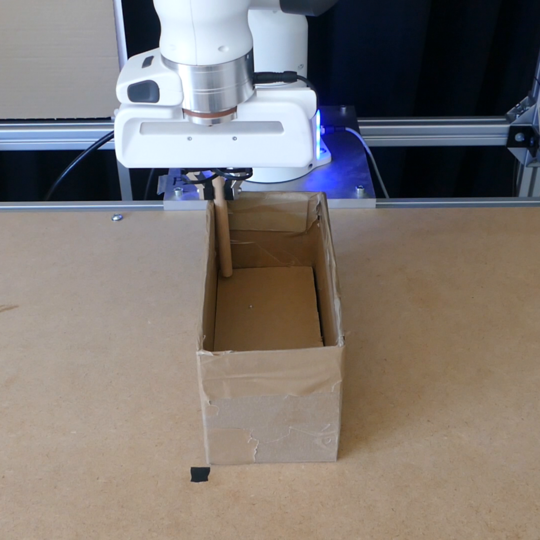}
	\end{minipage}\hfill
	\begin{minipage}[t!]{0.138\textwidth}
		\includegraphics[width=\textwidth]{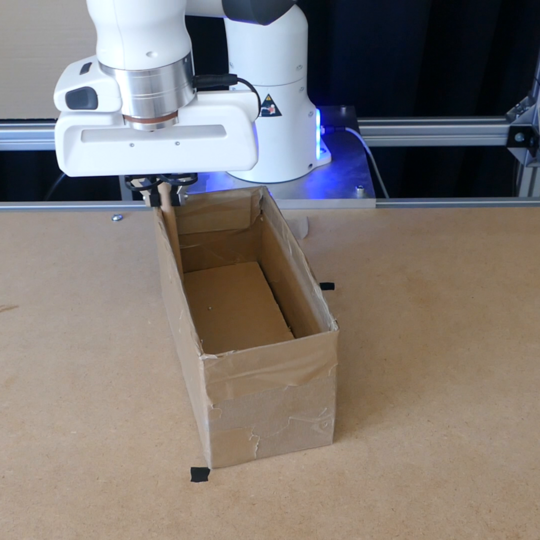}
	\end{minipage}\hfill
	\begin{minipage}[t!]{0.138\textwidth}
		\includegraphics[width=\textwidth]{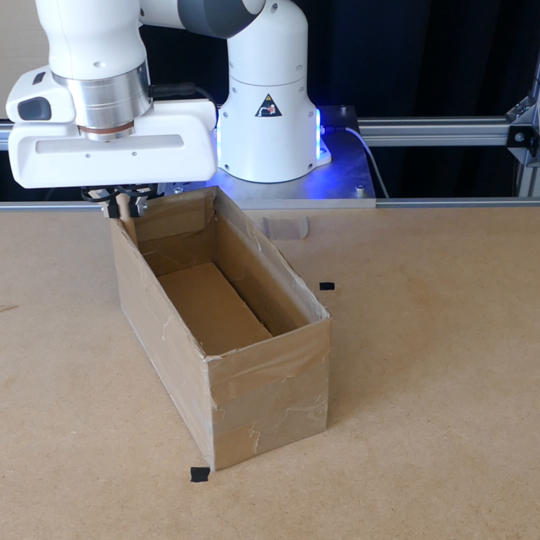}
	\end{minipage}\hfill
	\begin{minipage}[t!]{0.138\textwidth}
		\includegraphics[width=\textwidth]{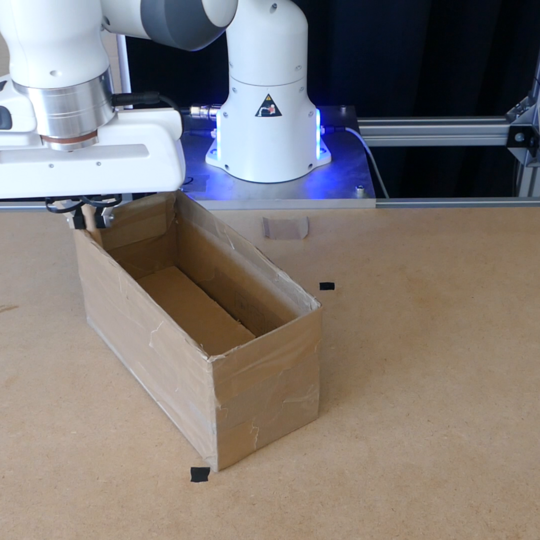}
	\end{minipage}\hfill
	\begin{minipage}[t!]{0.138\textwidth}
		\includegraphics[width=\textwidth]{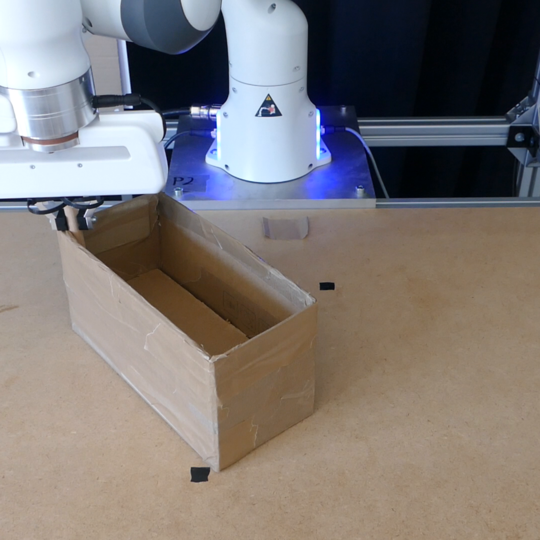}
	\end{minipage}\hfill
  \end{minipage}\hfill
  \begin{minipage}[t!]{\textwidth}
    \vspace{0.4mm}
	\begin{minipage}[t!]{0.138\textwidth}
		\includegraphics[width=\textwidth]{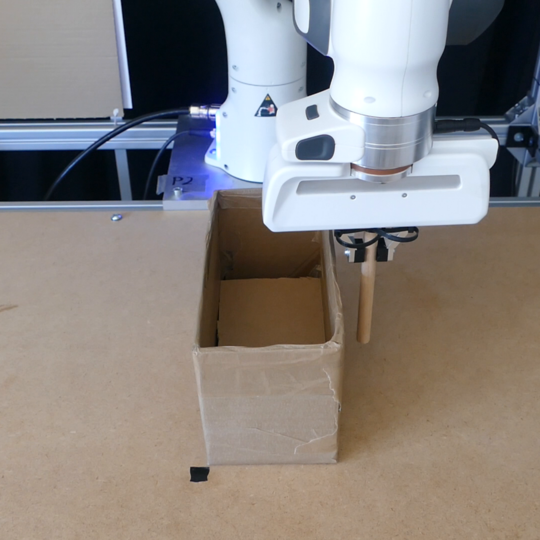}
	\end{minipage}\hfill
	\begin{minipage}[t!]{0.138\textwidth}
		\includegraphics[width=\textwidth]{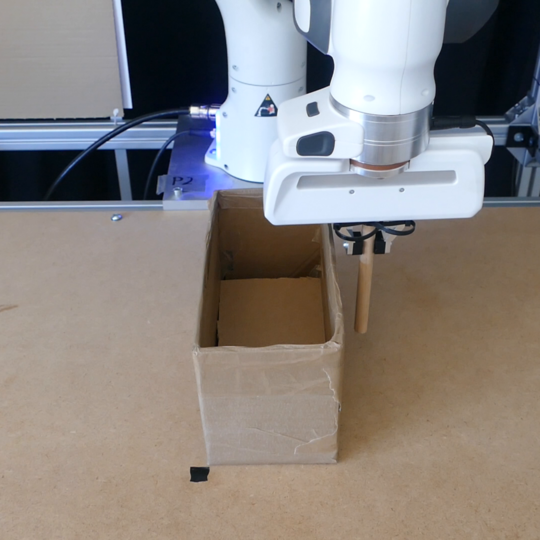}
	\end{minipage}\hfill
	\begin{minipage}[t!]{0.138\textwidth}
		\includegraphics[width=\textwidth]{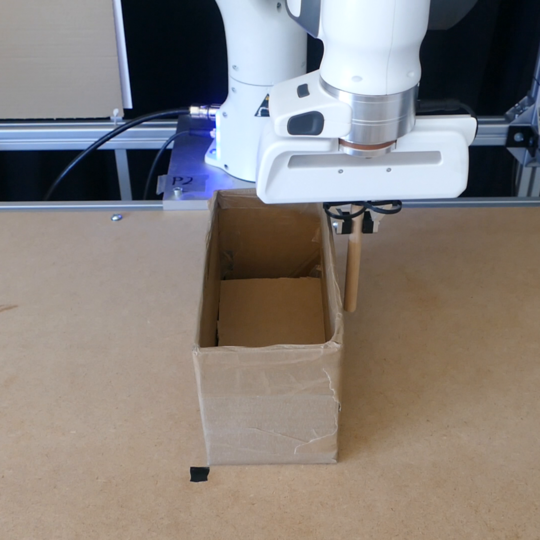}
	\end{minipage}\hfill
	\begin{minipage}[t!]{0.138\textwidth}
		\includegraphics[width=\textwidth]{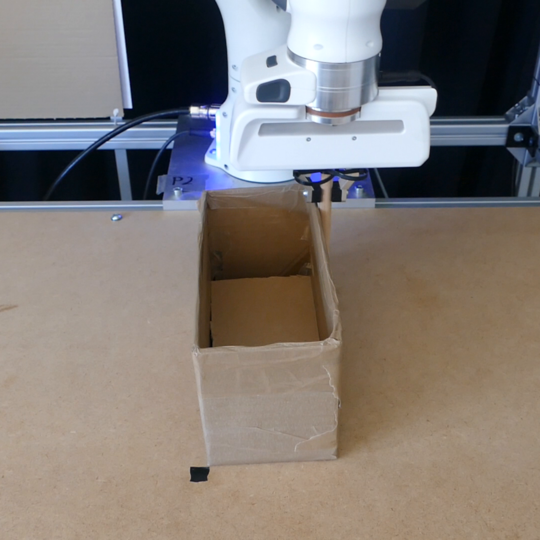}
	\end{minipage}\hfill
	\begin{minipage}[t!]{0.138\textwidth}
		\includegraphics[width=\textwidth]{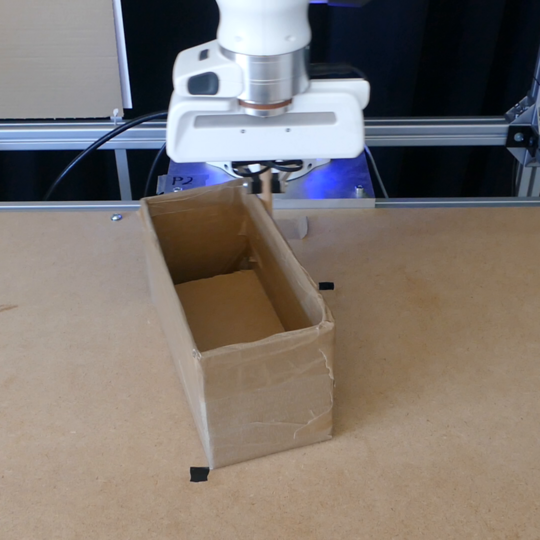}
	\end{minipage}\hfill
	\begin{minipage}[t!]{0.138\textwidth}
		\includegraphics[width=\textwidth]{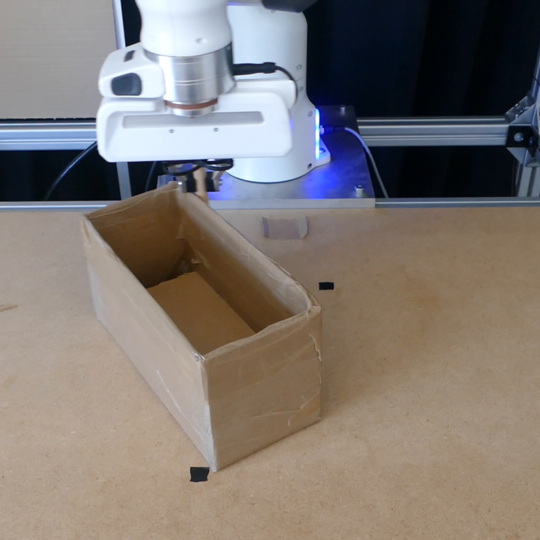}
	\end{minipage}\hfill
	\begin{minipage}[t!]{0.138\textwidth}
		\includegraphics[width=\textwidth]{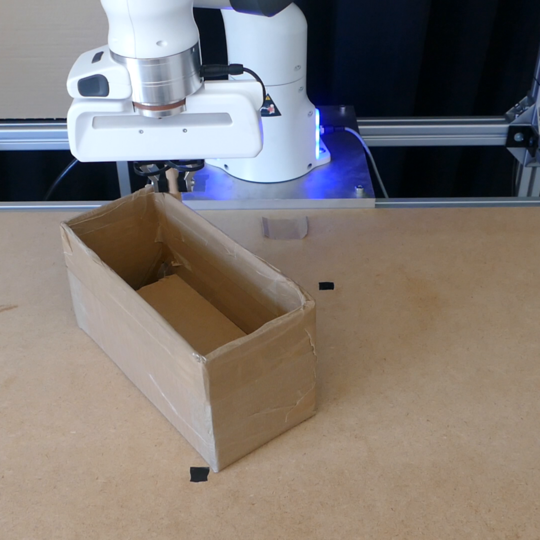}
	\end{minipage}\hfill
  \end{minipage}\hfill
	
  \begin{minipage}[t!]{\textwidth}
    \vspace{0.4mm}
	\begin{minipage}[t!]{0.138\textwidth}
		\includegraphics[width=\textwidth]{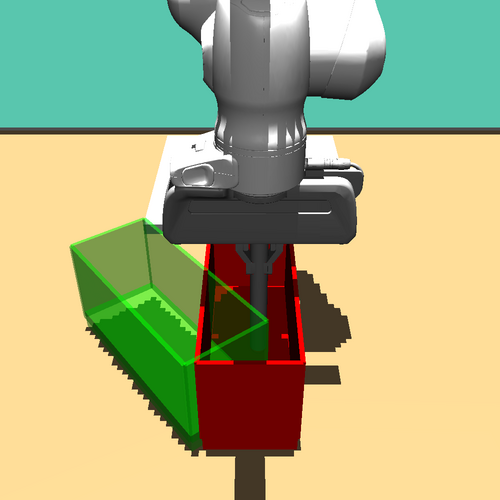}
	\end{minipage}\hfill
	\begin{minipage}[t!]{0.138\textwidth}
		\includegraphics[width=\textwidth]{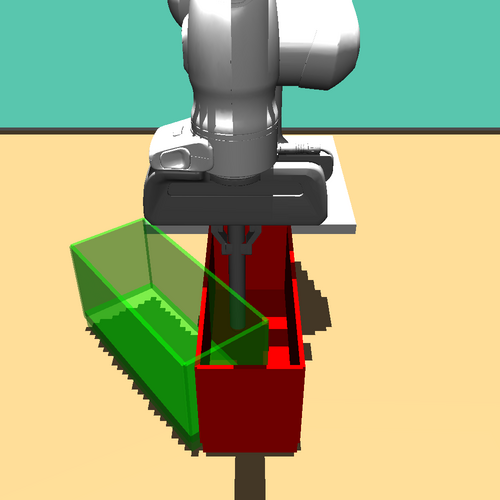}
	\end{minipage}\hfill
	\begin{minipage}[t!]{0.138\textwidth}
		\includegraphics[width=\textwidth]{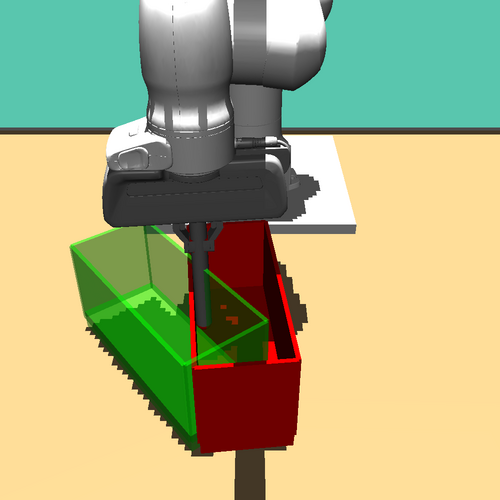}
	\end{minipage}\hfill
	\begin{minipage}[t!]{0.138\textwidth}
		\includegraphics[width=\textwidth]{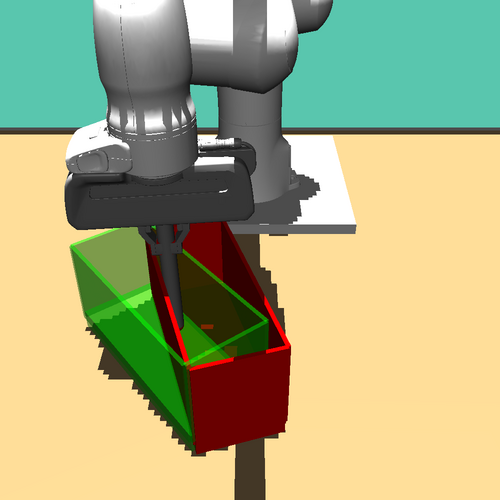}
	\end{minipage}\hfill
	\begin{minipage}[t!]{0.138\textwidth}
		\includegraphics[width=\textwidth]{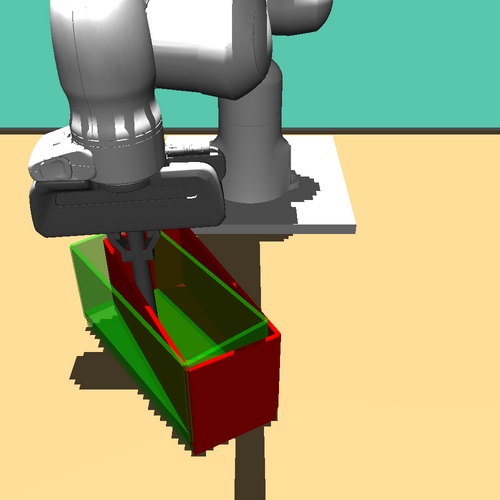}
	\end{minipage}\hfill
	\begin{minipage}[t!]{0.138\textwidth}
		\includegraphics[width=\textwidth]{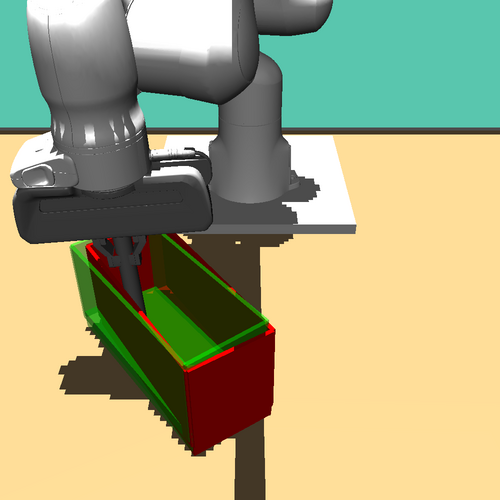}
	\end{minipage}\hfill
	\begin{minipage}[t!]{0.138\textwidth}
		\includegraphics[width=\textwidth]{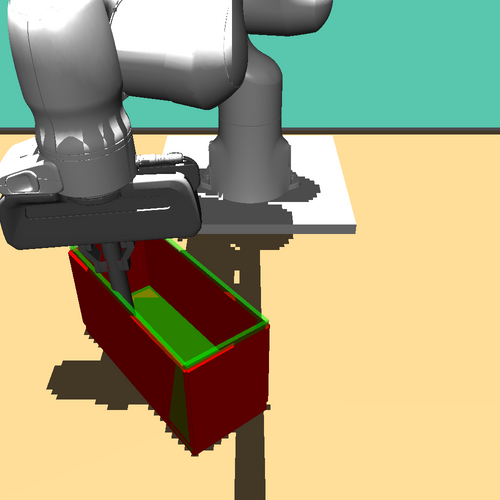}
	\end{minipage}\hfill

  \end{minipage}\hfill
\end{minipage}\hfill
	\vspace{-2mm}
   \caption{
    Trajectory executions of three component means of a \gls{vigor} policy trained on $24$ training contexts for the same test context.
    The first two rows show a real Franka Panda executing the planned end-effector trajectories. 
    The last row shows the same setup in the simulation, where the target box is overlaid in green.
    All three executions push the box in different ways, capturing the versatility of the human demonstrations; while the robot in the first row pushes the box from the inside, the robot in the second row pushes it from the outside. 
    The third trajectory also pushes from the inside but is much farther away from the corner of the box.
   }
   \label{fig:box_pusher_real_robots}
   \vspace{-1.5mm}
\end{figure}

\textbf{Ablations.}\label{ssec:ablations}
We conduct extensive ablation studies to investigate how our design choices affect our approach. We find that a concise choice of behavioral descriptors, a preprocessing of the expert demonstrations to ProMPs, and a $1d$-CNN discriminator are crucial for performance. We also notice that the policy benefits from additional components and that we see modest improvements for an ensemble of discriminators and the stepwise loss proposed in Equation \ref{eq:stepwise_bce}.
We refer to Appendix \ref{app_sec:ablations} for the full results and more thorough ablations of both \gls{vigor} and \gls{em}+\gls{drex}.


\section{Limitations}
\label{sec:limitations}

\textbf{Scale.} 
We currently train and maintain a separate \gls{gmm} over \gls{promp} parameters for each context, leading to linear space and time complexity w.r.t. the total number of contexts. One way to address this challenge would be to instead train a joint Mixture Density Network $q(\w|c)$ over all contexts, which is left for future work.
Additionally, representing trajectories with a single \gls{promp} limits them to single smooth movements. To alleviate this, \gls{mp} \textit{chaining}~\citep{daniel2016hierarchical, neumann2009learning, manschitz2014learning} can be straightforwardly integrated into our approach to allow for the representation of more complex movements

\textbf{Geometric Descriptors.}
\gls{vigor} facilitates generalization to novel task configurations by using geometric descriptors.
While this allows the user to include domain knowledge, it also requires a clear idea about which aspects of the task are important.
Instead, geometric descriptors could be automatically extracted from the task for example by using NNs processing point clouds~\citep{qi2017pointnet, qi2017pointnet++}.

\textbf{Re-training.} 
Finally, our method assumes that the test configurations $\ctx_{\text{test}}$ are known in advance.
Since the test configurations are optimized jointly with the training configurations, the only way to integrate new test configurations is to re-train the algorithm. 
In the future, we want to alleviate this issue by recovering a reward representation from ranked intermediate policies of the given training configurations, essentially combining our approach with some of the ideas of \gls{drex}.

\section{Conclusion}
\label{sec:conclusion}
We proposed \gls{vigor}, a novel feature-matching approach to Imitation Learning in environments with versatile solutions. 
Utilizing a combination of movement primitives, mixture policies and matching geometric behavioral descriptors, our method can closely imitate the behavior distribution of human experts from a few demonstrations.
We show the effectiveness of our approach on a suite of challenging robot coordination tasks. The results show that \gls{vigor} is able to closely match the distribution of the demonstrator, outperforming the chosen baselines in all considered settings.

\clearpage

\acknowledgments{
NS and GN were supported by the Carl Zeiss Foundation under the project JuBot (Jung Bleiben mit Robotern).
The authors acknowledge support by the state of Baden-Württemberg through bwHPC.
}


\bibliography{bibliography}

\newpage
\appendix
\section{Pseudocode}
\label{app_sec:pseudocode}

Pseudocode for \gls{vigor} is provided in Algorithm \ref{alg:vigor}.

\setcounter{algocf}{-1} 
\begin{algorithm}[t]
\SetAlgoLined
\DontPrintSemicolon
\caption{\gls{vigor}}
\KwInput{Contexts $\ctx = \ctx_{\text{train}} \cup \ctx_{\text{test}}$}
\KwInput{Mappings $f_c$ with $\Obs=f_c(\w)$ for each $c\in \ctx$}
\KwInput{Expert Descriptors $\Obs^{(p)} = \{\Obs^{(p)}_c | c\in \ctx_{\text{train}}\}$}
\KwInput{Initial Policies $\{q_c(\w) | c\in \ctx\}$}
\KwOutput{Converged Policies $\{q_c(\w) | c\in \ctx\}$}
$n_{\text{samples}} \leftarrow |\Obs^{(p)}|/|\ctx|$\\
$\hat{q_c}(\w) \leftarrow q_c(\w) \quad \forall c\in\ctx$\\

\While{not converged}{
    \textbf{Gather new policy samples}\\
    $\Obs^{(q)} \leftarrow \{\}$\\
    \For{$c\in \ctx$}{ 
        $\hat{\Obs}^{(q)}_c \leftarrow \{\w_c^{\left(j\right)}\}_{j=1\dots n_{\text{samples}}} \sim q_c\left(\w\right)$ \\
        $\Obs^{(q)} \leftarrow \Obs^{(q)} \cup \{f_c\left(\w\right) | \w \in \hat{\Obs}^{(q)}_c \}$
    }
    ~\\
    \textbf{Train discriminator $\bphi(\Obs)$ (c.f. Eq. \ref{eq:stepwise_bce})}\\
    $\bphi(\Obs) \leftarrow \argmin_{\bphi(\Obs)}\text{BCE}(\bphi(\Obs), \Obs^{(p)}, \Obs^{(q)})$  
    \\
    ~\\
    \textbf{Update policies for each context with discriminator $\bphi(\Obs)$ (c.f. Eq. \ref{eq:vips_objective}, )}\\
    \For{$c\in \ctx$}{
        \begin{align*}
        q_c\left(\w\right) \leftarrow \argmin_{q\left(\w\right)} 
        &\mathbb{E}_{q\left(\w, \z\right)} 
        \left[ \bphi(\Obs(\w)) \right] + \text{KL}\big(q(\z) || \hat{q}_{c}(\z)\big) + \\  &\mathbb{E}_{q\left(\z\right)}\Big[\text{KL}\big(q\left(\w|\z\right) || \hat{q}_{c}\left(\w|\z\right)\big)\Big]
        \end{align*}
    
    $\hat{q}_{c}(\w|\z) \leftarrow q(\w|\z) \quad \forall \z $ \\
    }
}
\KwRet{$\{q_c(\w) | c\in \ctx\}$}
\caption{\gls{vigor}}
\label{alg:vigor}
\end{algorithm}

\section{Baselines}
\label{app_sec:baselines}
We compare \gls{vigor} to a number of strong \gls{il} baselines in both a trajectory-based and state-action setting.

\subsection{EM+D-REX} 
\label{app_ssec:em_drex}
We modify \glsfirst{drex}~\citep{brown2020better}, a state-of-the-art \gls{il} algorithm to work in a trajectory-based setting and with a \glspl{gmm} policy. 
For this, we first use \gls{em} to fit a \gls{gmm} with a small number of components on each configuration in $\ctx_{\text{train}}$, and use the resulting distributions to draw samples for the \gls{drex} reward. 
To generate the rankings required by \gls{drex}, we multiply the covariance of each component of the resulting \gls{gmm} with different scalars, resulting in \glspl{gmm} with different noise levels. More precisely, we fit the \gls{em}-\gls{gmm} on the samples, and multiply its component-wise covariance with a \textit{Base Noise} $\sigma_{\text{base}}$ to get an initial distribution. We repeat this process for a number of \textit{Noise Levels}, linearly increasing the noise up to $\sigma_{\text{base}}\cdot \lambda_{\text{max}}$, where we call $\lambda_{\text{max}}$ the \textit{Noise Multiplier}. 
We then draw samples from each noise level of the \glspl{gmm} corresponding to each context in $\ctx_{\text{train}}$ and rank them against each other using the comparison-based loss of \gls{drex}, where samples drawn from a lower noise level are ranked higher.
The resulting ranked demonstrations are transformed into geometric descriptors $\Obs$ and used to train a reward function $R(\Obs)$, which in turn is used by a policy optimization algorithm to optimize policies on unknown contexts $\ctx_{\text{test}}$.
For the policy optimization algorithm, we use \gls{vips}~\citep{arenz2018efficient, arenz2020trust}. 
To make results more stable, we optimize on a scaled reward $\hat{R}(\Obs)=\alpha_{\text{scale}}R(\Obs)$ instead of the regular reward, where $\alpha_{\text{scale}}$ is a scalar \textit{Reward Scale}. We call the resulting algorithm \gls{em}+\gls{drex}.
Comparing EM+D-REX and \gls{vigor}, both approaches train a discriminator that uses expert demonstrations on training contexts to fit novel test contexts. 
Similarly, both make use of \gls{vips} to optimize marginal policies to match multi-modal distributions over geometric behavioral descriptors.
However, EM+D-REX uses the discriminator to represent a reward function, while VIGOR iteratively re-trains the discriminator until convergence of the test policies.

\subsection{State-Action Baselines} 
We also compare our approach to common state-action based imitation learning algorithms to see how well these approaches work on versatile human demonstrations. More precisely, we compare to \gls{bc}~\citep{bain1995framework} and \glsfirst{gail}~\citep{ho2016generative} as implemented in the \textit{Imitation}~\citep{wang2020imitation} repository. 
The state-action baselines use velocities as actions, and receive the geometric descriptors of the current robot state plus the current time-step as state information.
For the Point Reaching tasks, we also experiment with encoding which of the points has been reached to make the tasks Markovian.
We use the \gls{ppo}~\citep{schulman2017proximal} implementation of Stable Baselines3~\citep{stable-baselines3} as the policy for \gls{gail}. 
To accommodate for versatility on an action level, we also compare to BC-GMM~\citep{mandlekar2021matters}, which trains a policy whose head outputs the parameters of a GMM over actions per state. We note that we do not make use of the Low Noise Evaluation Trick proposed in \citet{mandlekar2021matters}, which leads to minor improvements in their experiments and does not affect our results. 
During evaluation, we also generate each trajectory from a fixed component rather than sampling a new component mean per step for consistency with the other approaches.
Experimentally, we found no significant difference in performance between these two evaluation methods.
In our experiments, both \gls{bc} and BC-GMM also make use of an auxiliary entropy regularization term that is similar to \citet{zhou2020movement}. 
This prevents the covariances from collapsing.
All state-action baselines use regular MLPs for their policy.

\subsection{Trajectory-based Baselines}
Finally, we employ \gls{bc} and BC-GMM on a trajectory-based level, i.e., we imitate full trajectories instead of individual state-action pairs. 
These baselines are added to see how simple (multi-modal) behavioral cloning works on full expert trajectories. 
In this setup, the methods see as input the context of the task, e.g., the position of the target points for the point reaching tasks, and output a contextual distribution $q(\w|c)$ over \gls{promp} parameters. 
This distribution is implemented as a simple MLP.
To discriminate between  state-action and trajectory-based baselines, we append suffixes `(S)' and `(T)' respectively. For example, we denote state-action BC as `BC(S)' and trajectory-based Mixture-Density BC as `BC-GMM(T)'.

\section{Additional Task Information and Results}
\label{app_sec:task_info_and_results}

\subsection{Planar Reacher}
\label{app_sec:planar_point_reacher}

\paragraph{Setup.} 
Each context is specified by its target positions, which are drawn from independent isotropic Gaussians with means $(0.5, 2.5)$ and $(0.5, -2.5)$ and standard deviation $0.5$.
Both targets have a radius of $r=0.5$ to be considered \textit{reached}.
The robot always starts with all joints at a resting position of $0\deg$.
We collect demonstrations via a joystick-based setup to control the end-effector's $(x,y)$-position and rotation. This is paired with an inverse kinematics controller to control the joints.

\begin{figure}[t]
    \centering
    \begin{minipage}{0.5\textwidth}
        \resizebox{\textwidth}{!}{
\begin{tikzpicture}

\definecolor{brown1926061}{RGB}{192,60,61}
\definecolor{darkgray176}{RGB}{176,176,176}
\definecolor{darkslategray61}{RGB}{61,61,61}
\definecolor{dimgray1319183}{RGB}{131,91,83}
\definecolor{forestgreen4416044}{RGB}{0,0,0}
\definecolor{lightgray204}{RGB}{204,204,204}
\definecolor{mediumpurple147113178}{RGB}{147,113,178}
\definecolor{orchid213132188}{RGB}{213,132,188}
\definecolor{peru22412844}{RGB}{224,128,44}
\definecolor{seagreen5814558}{RGB}{58,145,58}
\definecolor{steelblue49115161}{RGB}{49,115,161}

\begin{axis}[
legend style={
  fill opacity=0.8,
  draw opacity=1,
  text opacity=1,
  at={(0.03,0.97)},
  anchor=north west,
  draw=lightgray204
},
tick align=outside,
tick pos=left,
x grid style={darkgray176},
xmajorgrids,
xmin=-0.5, xmax=6.5,
xtick style={color=black},
xtick={0,1,2,3,4,5,6},
xticklabels={VIGOR,EM+\\D-REX,BC-\\GMM(T),BC(T),GAIL,BC-\\GMM(S),BC(S)},
y grid style={darkgray176},
ylabel={Success Rate},
ymajorgrids,
ymin=-0.1, ymax=1.05,
ytick style={color=black},
height=5.5cm, 
width=8cm, 
x tick label style={rotate=90, font=\small, align=right}
]
\path [draw=darkslategray61, fill=steelblue49115161, semithick]
(axis cs:-0.4,0.993333339691162)
--(axis cs:0.4,0.993333339691162)
--(axis cs:0.4,1)
--(axis cs:-0.4,1)
--(axis cs:-0.4,0.993333339691162)
--cycle;
\path [draw=darkslategray61, fill=peru22412844, semithick]
(axis cs:0.6,0.673750013113022)
--(axis cs:1.4,0.673750013113022)
--(axis cs:1.4,1)
--(axis cs:0.6,1)
--(axis cs:0.6,0.673750013113022)
--cycle;
\path [draw=darkslategray61, fill=seagreen5814558, semithick]
(axis cs:1.6,0.0283333328552544)
--(axis cs:2.4,0.0283333328552544)
--(axis cs:2.4,0.143333336648842)
--(axis cs:1.6,0.143333336648842)
--(axis cs:1.6,0.0283333328552544)
--cycle;
\path [draw=darkslategray61, fill=brown1926061, semithick]
(axis cs:2.6,0)
--(axis cs:3.4,0)
--(axis cs:3.4,0.00333333318121731)
--(axis cs:2.6,0.00333333318121731)
--(axis cs:2.6,0)
--cycle;
\path [draw=darkslategray61, fill=mediumpurple147113178, semithick]
(axis cs:3.6,0)
--(axis cs:4.4,0)
--(axis cs:4.4,0.0341666666666667)
--(axis cs:3.6,0.0341666666666667)
--(axis cs:3.6,0)
--cycle;
\path [draw=darkslategray61, fill=dimgray1319183, semithick]
(axis cs:4.6,0.0025)
--(axis cs:5.4,0.0025)
--(axis cs:5.4,0.01625)
--(axis cs:4.6,0.01625)
--(axis cs:4.6,0.0025)
--cycle;
\path [draw=darkslategray61, fill=orchid213132188, semithick]
(axis cs:5.6,0.00333333333333333)
--(axis cs:6.4,0.00333333333333333)
--(axis cs:6.4,0.00625)
--(axis cs:5.6,0.00625)
--(axis cs:5.6,0.00333333333333333)
--cycle;
\addplot [semithick, darkslategray61, forget plot]
table {%
0 0.993333339691162
0 0.991666615009308
};
\addplot [semithick, darkslategray61, forget plot]
table {%
0 1
0 1
};
\addplot [semithick, darkslategray61, forget plot]
table {%
-0.2 0.991666615009308
0.2 0.991666615009308
};
\addplot [semithick, darkslategray61, forget plot]
table {%
-0.2 1
0.2 1
};
\addplot [black, mark=diamond*, mark size=2.5, mark options={solid,fill=darkslategray61}, only marks, forget plot]
table {%
0 0.936666667461395
};
\addplot [semithick, darkslategray61, forget plot]
table {%
1 0.673750013113022
1 0.516666650772095
};
\addplot [semithick, darkslategray61, forget plot]
table {%
1 1
1 1
};
\addplot [semithick, darkslategray61, forget plot]
table {%
0.8 0.516666650772095
1.2 0.516666650772095
};
\addplot [semithick, darkslategray61, forget plot]
table {%
0.8 1
1.2 1
};
\addplot [semithick, darkslategray61, forget plot]
table {%
2 0.0283333328552544
2 0.00499999988824129
};
\addplot [semithick, darkslategray61, forget plot]
table {%
2 0.143333336648842
2 0.173333334426085
};
\addplot [semithick, darkslategray61, forget plot]
table {%
1.8 0.00499999988824129
2.2 0.00499999988824129
};
\addplot [semithick, darkslategray61, forget plot]
table {%
1.8 0.173333334426085
2.2 0.173333334426085
};
\addplot [semithick, darkslategray61, forget plot]
table {%
3 0
3 0
};
\addplot [semithick, darkslategray61, forget plot]
table {%
3 0.00333333318121731
3 0.00666666636243463
};
\addplot [semithick, darkslategray61, forget plot]
table {%
2.8 0
3.2 0
};
\addplot [semithick, darkslategray61, forget plot]
table {%
2.8 0.00666666636243463
3.2 0.00666666636243463
};
\addplot [semithick, darkslategray61, forget plot]
table {%
4 0
4 0
};
\addplot [semithick, darkslategray61, forget plot]
table {%
4 0.0341666666666667
4 0.0816666666666667
};
\addplot [semithick, darkslategray61, forget plot]
table {%
3.8 0
4.2 0
};
\addplot [semithick, darkslategray61, forget plot]
table {%
3.8 0.0816666666666667
4.2 0.0816666666666667
};
\addplot [semithick, darkslategray61, forget plot]
table {%
5 0.0025
5 0
};
\addplot [semithick, darkslategray61, forget plot]
table {%
5 0.01625
5 0.0283333333333333
};
\addplot [semithick, darkslategray61, forget plot]
table {%
4.8 0
5.2 0
};
\addplot [semithick, darkslategray61, forget plot]
table {%
4.8 0.0283333333333333
5.2 0.0283333333333333
};
\addplot [black, mark=diamond*, mark size=2.5, mark options={solid,fill=darkslategray61}, only marks, forget plot]
table {%
5 0.05
};
\addplot [semithick, darkslategray61, forget plot]
table {%
6 0.00333333333333333
6 0
};
\addplot [semithick, darkslategray61, forget plot]
table {%
6 0.00625
6 0.00833333333333333
};
\addplot [semithick, darkslategray61, forget plot]
table {%
5.8 0
6.2 0
};
\addplot [semithick, darkslategray61, forget plot]
table {%
5.8 0.00833333333333333
6.2 0.00833333333333333
};
\addplot [semithick, darkslategray61, forget plot]
table {%
-0.4 0.998333334922791
0.4 0.998333334922791
};
\addplot [forestgreen4416044, mark=x, mark size=3, mark options={solid,fill=black}, only marks, forget plot]
table {%
0 0.991166663169861
};
\addplot [semithick, darkslategray61, forget plot]
table {%
0.6 0.871666669845581
1.4 0.871666669845581
};
\addplot [forestgreen4416044, mark=x, mark size=3, mark options={solid,fill=black}, only marks, forget plot]
table {%
1 0.828333336114883
};
\addplot [semithick, darkslategray61, forget plot]
table {%
1.6 0.0775000001303852
2.4 0.0775000001303852
};
\addplot [forestgreen4416044, mark=x, mark size=3, mark options={solid,fill=black}, only marks, forget plot]
table {%
2 0.0846666669162611
};
\addplot [semithick, darkslategray61, forget plot]
table {%
2.6 0.00166666659060866
3.4 0.00166666659060866
};
\addplot [forestgreen4416044, mark=x, mark size=3, mark options={solid,fill=black}, only marks, forget plot]
table {%
3 0.00216666657943279
};
\addplot [semithick, darkslategray61, forget plot]
table {%
3.6 0
4.4 0
};
\addplot [forestgreen4416044, mark=x, mark size=3, mark options={solid,fill=black}, only marks, forget plot]
table {%
4 0.0208333333333333
};
\addplot [semithick, darkslategray61, forget plot]
table {%
4.6 0.0125
5.4 0.0125
};
\addplot [forestgreen4416044, mark=x, mark size=3, mark options={solid,fill=black}, only marks, forget plot]
table {%
5 0.0145
};
\addplot [semithick, darkslategray61, forget plot]
table {%
5.6 0.005
6.4 0.005
};
\addplot [forestgreen4416044, mark=x, mark size=3, mark options={solid,fill=black}, only marks, forget plot]
table {%
6 0.0045
};
\end{axis}

\end{tikzpicture}}
    \end{minipage}%
    \begin{minipage}{0.5\textwidth}
        \resizebox{\textwidth}{!}{
\begin{tikzpicture}

\definecolor{brown1926061}{RGB}{192,60,61}
\definecolor{darkgray176}{RGB}{176,176,176}
\definecolor{darkslategray61}{RGB}{61,61,61}
\definecolor{dimgray1319183}{RGB}{131,91,83}
\definecolor{forestgreen4416044}{RGB}{0,0,0}
\definecolor{lightgray204}{RGB}{204,204,204}
\definecolor{mediumpurple147113178}{RGB}{147,113,178}
\definecolor{orchid213132188}{RGB}{213,132,188}
\definecolor{peru22412844}{RGB}{224,128,44}
\definecolor{seagreen5814558}{RGB}{58,145,58}
\definecolor{steelblue49115161}{RGB}{49,115,161}

\begin{axis}[
legend style={
  fill opacity=0.8,
  draw opacity=1,
  text opacity=1,
  at={(0.03,0.97)},
  anchor=north west,
  draw=lightgray204
},
tick align=outside,
tick pos=left,
x grid style={darkgray176},
xmajorgrids,
xmin=-0.5, xmax=6.5,
xtick style={color=black},
xtick={0,1,2,3,4,5,6},
xticklabels={VIGOR,EM+\\D-REX,BC-\\GMM(T),BC(T),GAIL,BC-\\GMM(S),BC(S)},
y grid style={darkgray176},
ylabel={Success Rate},
ymajorgrids,
ymin=-0.1, ymax=1.05,
ytick style={color=black},
height=5.5cm, 
width=8cm, 
x tick label style={rotate=90, font=\small, align=right}
]
\path [draw=darkslategray61, fill=steelblue49115161, semithick]
(axis cs:-0.4,1)
--(axis cs:0.4,1)
--(axis cs:0.4,1)
--(axis cs:-0.4,1)
--(axis cs:-0.4,1)
--cycle;
\path [draw=darkslategray61, fill=peru22412844, semithick]
(axis cs:0.6,0.514583326876163)
--(axis cs:1.4,0.514583326876163)
--(axis cs:1.4,0.992916658520699)
--(axis cs:0.6,0.992916658520699)
--(axis cs:0.6,0.514583326876163)
--cycle;
\path [draw=darkslategray61, fill=seagreen5814558, semithick]
(axis cs:1.6,0.000416666657353441)
--(axis cs:2.4,0.000416666657353441)
--(axis cs:2.4,0.00999999977648258)
--(axis cs:1.6,0.00999999977648258)
--(axis cs:1.6,0.000416666657353441)
--cycle;
\path [draw=darkslategray61, fill=brown1926061, semithick]
(axis cs:2.6,0)
--(axis cs:3.4,0)
--(axis cs:3.4,0)
--(axis cs:2.6,0)
--(axis cs:2.6,0)
--cycle;
\path [draw=darkslategray61, fill=mediumpurple147113178, semithick]
(axis cs:3.6,0.00166666666666667)
--(axis cs:4.4,0.00166666666666667)
--(axis cs:4.4,0.0158333333333333)
--(axis cs:3.6,0.0158333333333333)
--(axis cs:3.6,0.00166666666666667)
--cycle;
\path [draw=darkslategray61, fill=dimgray1319183, semithick]
(axis cs:4.6,0)
--(axis cs:5.4,0)
--(axis cs:5.4,0)
--(axis cs:4.6,0)
--(axis cs:4.6,0)
--cycle;
\path [draw=darkslategray61, fill=orchid213132188, semithick]
(axis cs:5.6,0)
--(axis cs:6.4,0)
--(axis cs:6.4,0)
--(axis cs:5.6,0)
--(axis cs:5.6,0)
--cycle;
\addplot [semithick, darkslategray61, forget plot]
table {%
0 1
0 1
};
\addplot [semithick, darkslategray61, forget plot]
table {%
0 1
0 1
};
\addplot [semithick, darkslategray61, forget plot]
table {%
-0.2 1
0.2 1
};
\addplot [semithick, darkslategray61, forget plot]
table {%
-0.2 1
0.2 1
};
\addplot [black, mark=diamond*, mark size=2.5, mark options={solid,fill=darkslategray61}, only marks, forget plot]
table {%
0 0.964999973773956
};
\addplot [semithick, darkslategray61, forget plot]
table {%
1 0.514583326876163
1 0.131666660308838
};
\addplot [semithick, darkslategray61, forget plot]
table {%
1 0.992916658520699
1 1
};
\addplot [semithick, darkslategray61, forget plot]
table {%
0.8 0.131666660308838
1.2 0.131666660308838
};
\addplot [semithick, darkslategray61, forget plot]
table {%
0.8 1
1.2 1
};
\addplot [semithick, darkslategray61, forget plot]
table {%
2 0.000416666657353441
2 0
};
\addplot [semithick, darkslategray61, forget plot]
table {%
2 0.00999999977648258
2 0.0133333330353101
};
\addplot [semithick, darkslategray61, forget plot]
table {%
1.8 0
2.2 0
};
\addplot [semithick, darkslategray61, forget plot]
table {%
1.8 0.0133333330353101
2.2 0.0133333330353101
};
\addplot [black, mark=diamond*, mark size=2.5, mark options={solid,fill=darkslategray61}, only marks, forget plot]
table {%
2 0.0500000007450581
};
\addplot [semithick, darkslategray61, forget plot]
table {%
3 0
3 0
};
\addplot [semithick, darkslategray61, forget plot]
table {%
3 0
3 0
};
\addplot [semithick, darkslategray61, forget plot]
table {%
2.8 0
3.2 0
};
\addplot [semithick, darkslategray61, forget plot]
table {%
2.8 0
3.2 0
};
\addplot [black, mark=diamond*, mark size=2.5, mark options={solid,fill=darkslategray61}, only marks, forget plot]
table {%
3 0.00166666659060866
3 0.00166666659060866
};
\addplot [semithick, darkslategray61, forget plot]
table {%
4 0.00166666666666667
4 0
};
\addplot [semithick, darkslategray61, forget plot]
table {%
4 0.0158333333333333
4 0.035
};
\addplot [semithick, darkslategray61, forget plot]
table {%
3.8 0
4.2 0
};
\addplot [semithick, darkslategray61, forget plot]
table {%
3.8 0.035
4.2 0.035
};
\addplot [semithick, darkslategray61, forget plot]
table {%
5 0
5 0
};
\addplot [semithick, darkslategray61, forget plot]
table {%
5 0
5 0
};
\addplot [semithick, darkslategray61, forget plot]
table {%
4.8 0
5.2 0
};
\addplot [semithick, darkslategray61, forget plot]
table {%
4.8 0
5.2 0
};
\addplot [semithick, darkslategray61, forget plot]
table {%
6 0
6 0
};
\addplot [semithick, darkslategray61, forget plot]
table {%
6 0
6 0
};
\addplot [semithick, darkslategray61, forget plot]
table {%
5.8 0
6.2 0
};
\addplot [semithick, darkslategray61, forget plot]
table {%
5.8 0
6.2 0
};
\addplot [semithick, darkslategray61, forget plot]
table {%
-0.4 1
0.4 1
};
\addplot [forestgreen4416044, mark=x, mark size=3, mark options={solid,fill=black}, only marks, forget plot]
table {%
0 0.996499997377396
};
\addplot [semithick, darkslategray61, forget plot]
table {%
0.6 0.861666679382324
1.4 0.861666679382324
};
\addplot [forestgreen4416044, mark=x, mark size=3, mark options={solid,fill=black}, only marks, forget plot]
table {%
1 0.722833330929279
};
\addplot [semithick, darkslategray61, forget plot]
table {%
1.6 0.00583333320294817
2.4 0.00583333320294817
};
\addplot [forestgreen4416044, mark=x, mark size=3, mark options={solid,fill=black}, only marks, forget plot]
table {%
2 0.00966666663686435
};
\addplot [semithick, darkslategray61, forget plot]
table {%
2.6 0
3.4 0
};
\addplot [forestgreen4416044, mark=x, mark size=3, mark options={solid,fill=black}, only marks, forget plot]
table {%
3 0.000333333318121731
};
\addplot [semithick, darkslategray61, forget plot]
table {%
3.6 0.0025
4.4 0.0025
};
\addplot [forestgreen4416044, mark=x, mark size=3, mark options={solid,fill=black}, only marks, forget plot]
table {%
4 0.00933333333333334
};
\addplot [semithick, darkslategray61, forget plot]
table {%
4.6 0
5.4 0
};
\addplot [forestgreen4416044, mark=x, mark size=3, mark options={solid,fill=black}, only marks, forget plot]
table {%
5 0
};
\addplot [semithick, darkslategray61, forget plot]
table {%
5.6 0
6.4 0
};
\addplot [forestgreen4416044, mark=x, mark size=3, mark options={solid,fill=black}, only marks, forget plot]
table {%
6 0
};
\end{axis}

\end{tikzpicture}}
    \end{minipage}
    \caption{
    (Left) Mean success rates on test contexts for the Planar Reacher task. 
    (Right) Mean success rates on test contexts for the Panda Reacher task. For both tasks, \gls{vigor} consistently reaches both target circles, while all other methods except EM+D-REX struggle to do so. In general, the baselines with a \gls{gmm} policy perform better than those without. GAIL performs the best of any state-action based approach.
    }
    \label{app_fig:reacher_success_rates}
\end{figure}
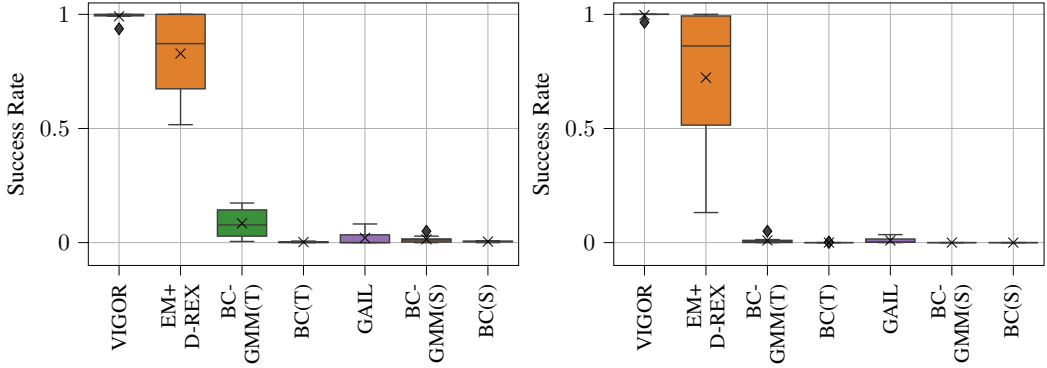

\paragraph{Success Rates.} 
We define a demonstration as successful if it reaches both target circles and ends its trajectory in the second circle, i.e., if its \textit{target distance} is $0$. 
As such, a demonstration is considered successful if it has the same target distance as the expert demonstrations. The left of Figure \ref{app_fig:reacher_success_rates} shows success rates on test contexts for different approaches trained on $6$ training contexts.

\begin{figure}[t]
\centering
VIGOR
\includegraphics[width=\textwidth]{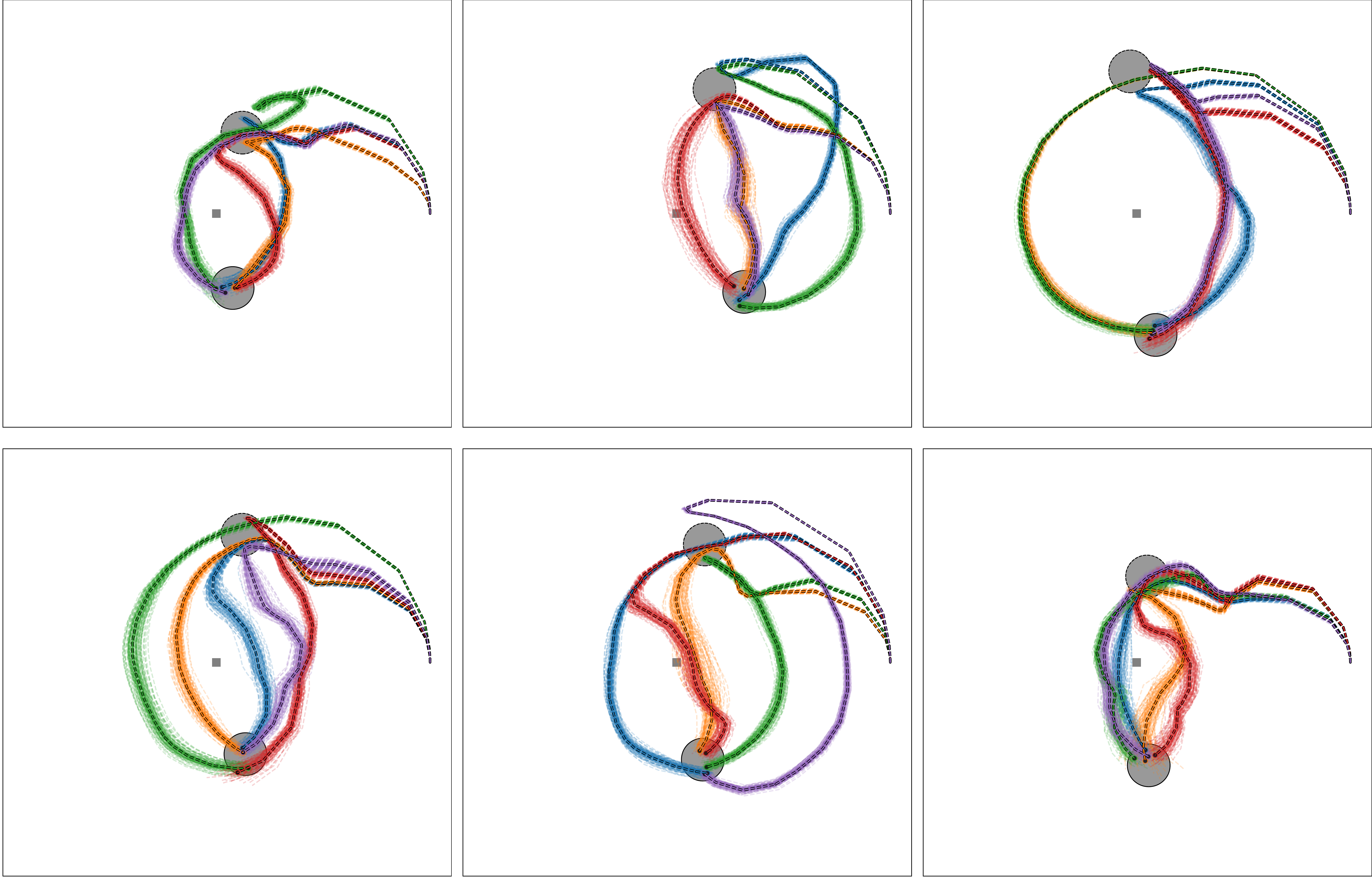}
EM+D-REX 
\includegraphics[width=\textwidth]{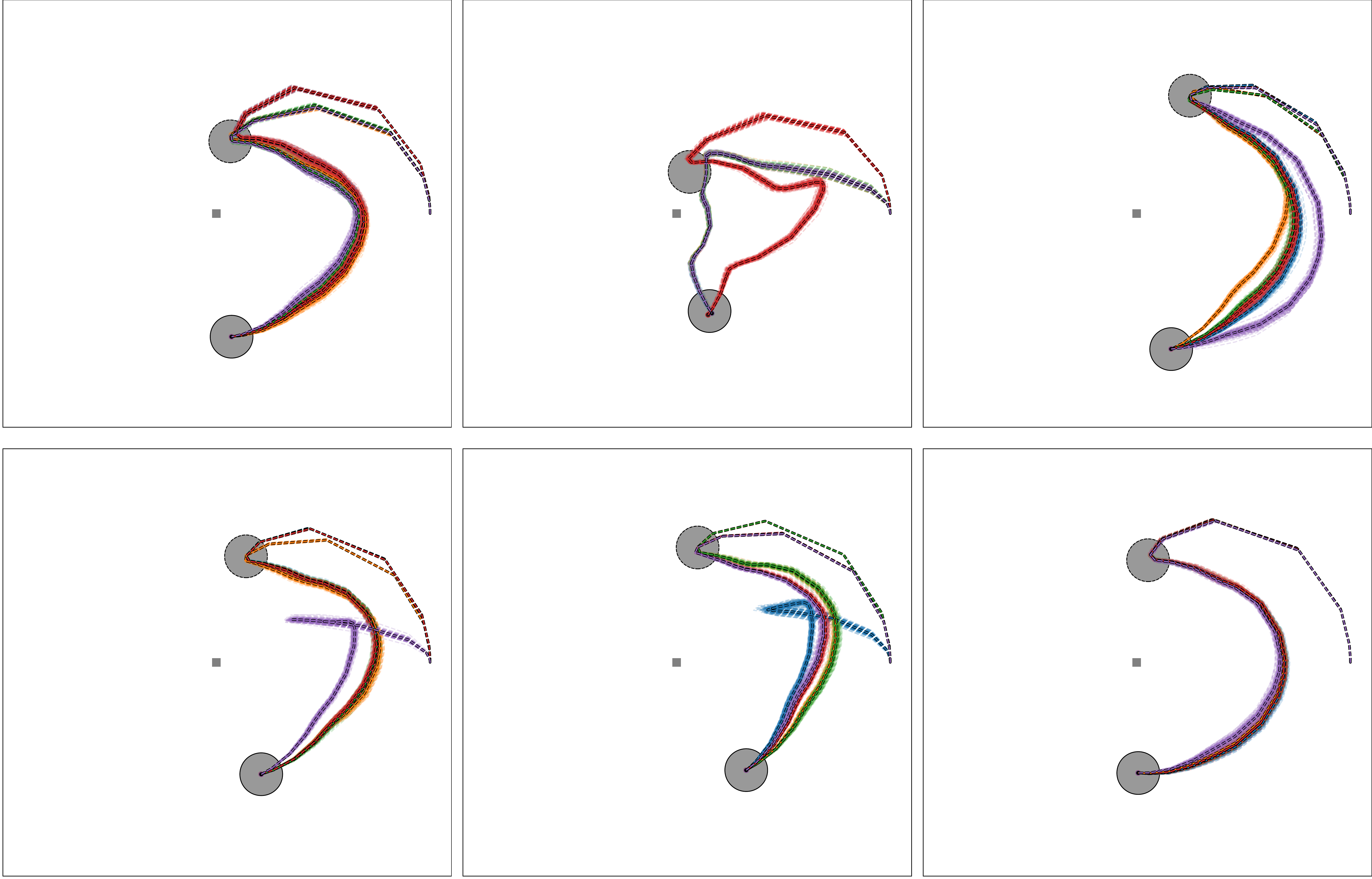}
    \caption{Visualization of end-effector traces of rollouts sampled from policies trained with \gls{vigor} and EM+D-REX. 
    Component means are marked with a black border, and for each component 100 samples are drawn using the same color. 
    Note how for most contexts, \gls{vigor} uses the components to specialize on different solutions, and how samples from this component are feasible variations of this solution.
    For EM+D-REX the components generally reach the targets but do not capture the versatility in the behavior.}
    \label{app_fig:reim_mpr_vis_samples}
\end{figure}

\paragraph{Qualitative results.} To explore the versatility of solutions learned by \gls{vigor} and \gls{em}+\gls{drex}, we visualize exemplary policies learned by both methods in Figures \ref{app_fig:reim_mpr_vis_samples}. 
We find that \gls{vigor} produces versatile policies that closely match the demonstrations of the expert, with most components finding a different solution to the given task and samples of these components generally hitting both targets. At the same time, the policies of \gls{em}+\gls{drex} often collapse to only $1$ or $2$ different solutions, likely due to a lack of a clear distribution matching objective. 

\subsection{Panda Reacher}
\label{app_sec:panda_reacher}
\begin{figure}[t]	
  \begin{minipage}[t!]{\textwidth}
  \begin{minipage}[t!]{\textwidth}
    \vspace{2mm}
	\begin{minipage}[t!]{0.32\textwidth}
		\includegraphics[width=\textwidth]{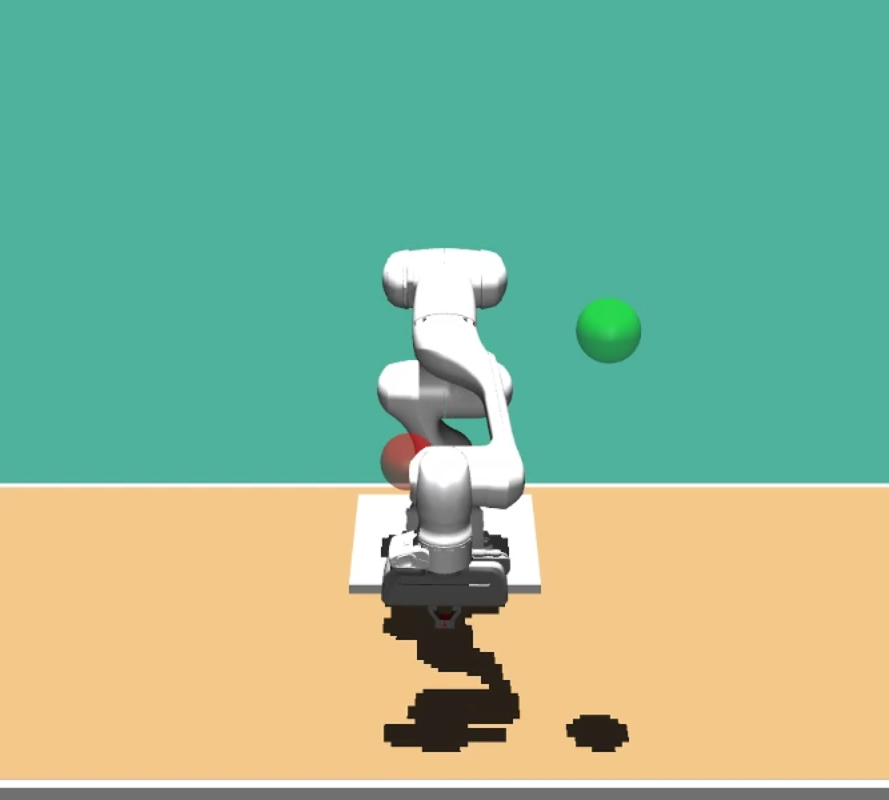}
	\end{minipage}\hfill
	\begin{minipage}[t!]{0.32\textwidth}
		\includegraphics[width=\textwidth]{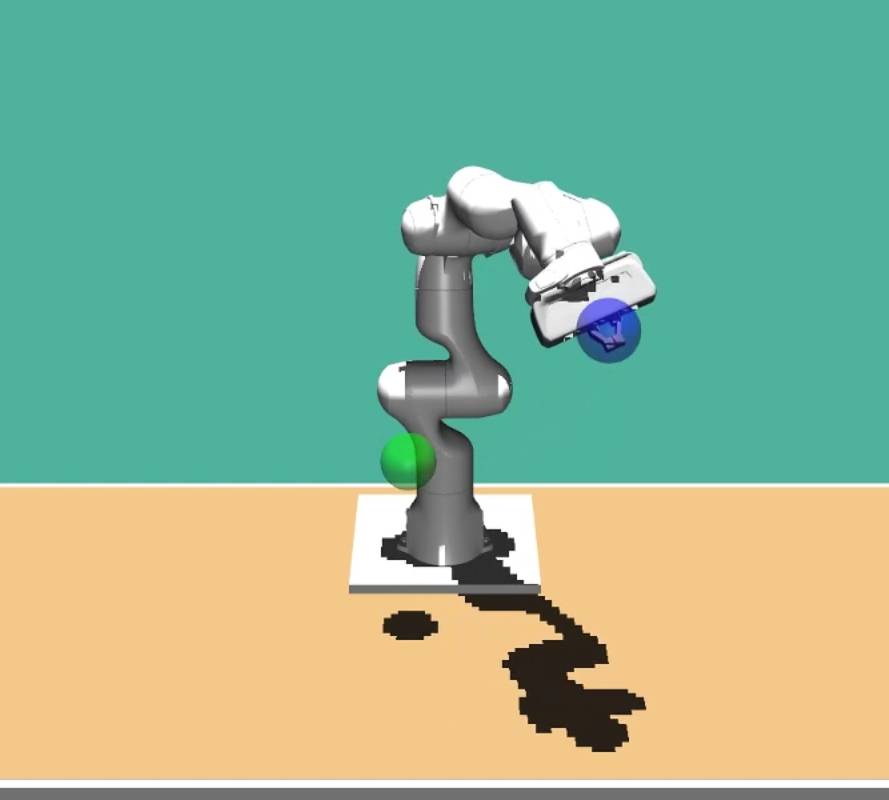}
	\end{minipage}\hfill
	\begin{minipage}[t!]{0.32\textwidth}
		\includegraphics[width=\textwidth]{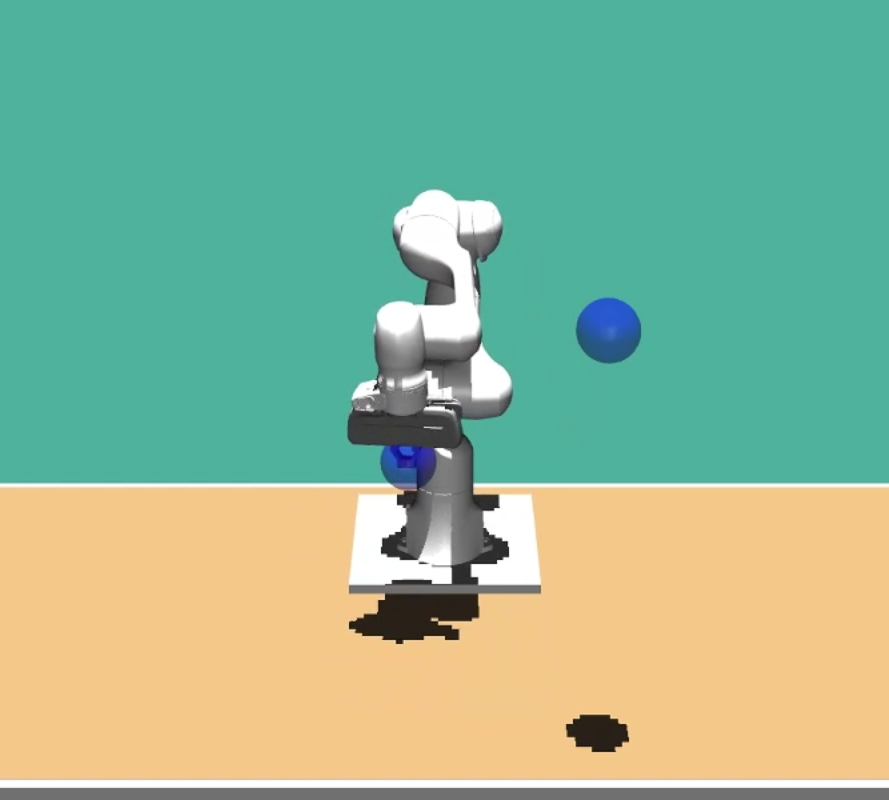}
	\end{minipage}\hfill
  \end{minipage}\hfill
\end{minipage}\hfill
	\vspace{-2mm}
   \caption{Virtual part of the setup for collecting human demonstrations for the Panda Reacher task. Starting from a resting position (left), the human demonstrator is tasked to reach the intermediate target (green). Once reached, both targets change color (middle). The task is successful if both targets have been reached (right).}
   \label{app_fig:teleoperation_twin_setup}
   \vspace{-1.5mm}
\end{figure}

\paragraph{Setup.} In the teleoperation setup, the human demonstrator moves a physical robot. The robot then sends its movement to a virtual twin. The joint values of the twin are recorded over time to generate expert demonstrations.
An overview of the virtual part of this setup is given in Figure \ref{app_fig:teleoperation_twin_setup}. 
The environment ranges from $(0, -0.5, 0)^T$ to $(1, 0.5, 1)^T$ meters.
The targets are drawn randomly from uniform distributions with range $0.1$ meters and means at $(0.5, 0.2, 0.6)^T$ and $(0.3, -0.1, 0.3)^T$ meters respectively. A target is considered reached within a radius of $0.05$ meters.
The robot starts all trajectories at its default resting position.

We preprocess the human demonstrations by removing initial steps until the robot starts moving, and repeating the last recorded time-step before fitting the \glspl{promp} to ensure that the fit trajectories end near the goal position. This is done to counteract a potential over-smoothing effect of the ProMPs fitting the human trajectories.

\paragraph{Success Rates.} The right of Figure \ref{app_fig:reacher_success_rates} shows success rates on test contexts for different approaches trained on $6$ training contexts. A demonstration is considered successful if it reaches both target circles and ends its trajectory in the second circle, i.e., if its \textit{target distance} is $0$cm. This corresponds to the target distance of the expert demonstrations.

\subsection{Box Pusher}
\label{app_sec:box_pusher}
\paragraph{Setup.} Demonstrations are collected using a virtual setup similar to that of the Planar Reacher task. The demonstrator controls a mouse in the $(x,y)$-plane that the end-effector of the robot follows using inverse kinematics. The rotation of the end-effector is fixed to have it point straight down. The height of the end-effector is fixed such that the rod is slightly above the table.
The trajectories, including the initial position of the end-effector, are fit in the $(x,y)$ plane of the task-space for simplicity. To roll out a given trajectory on the robot, the $(x,y)$ coordinates over time are concatenated with a fixed $z$-value and a downward rotation are given to the inverse kinematic controller.
We sample the contexts from box translations and rotations that are feasible to demonstrate within a single smooth movement. 
As there tends to be a high correlation between the $(x,y)$-displacement and the rotation angle in the resulting contexts, we make sure that the training contexts are balanced w.r.t. this correlation.

In order for the state-action baselines to be able to choose an initial position for the trajectory, we use a separate copy of the behavioral descriptors as input for the state-action at time-step $0$. 
In other words, we have $2$ inputs in the neural network for every feature. One of these is always $0$ except at step $0$, where it is set to the geometric descriptors. The other is always set to the geometric descriptors, except at step $0$, where it is set to $0$.
The output of the policy for this first time-step is then interpreted as the starting position relative to the center of the box. Outputs in later steps correspond to velocities in the $(x,y)$-plane.

\begin{figure}[t]
\centering
    \includegraphics[width=\textwidth]{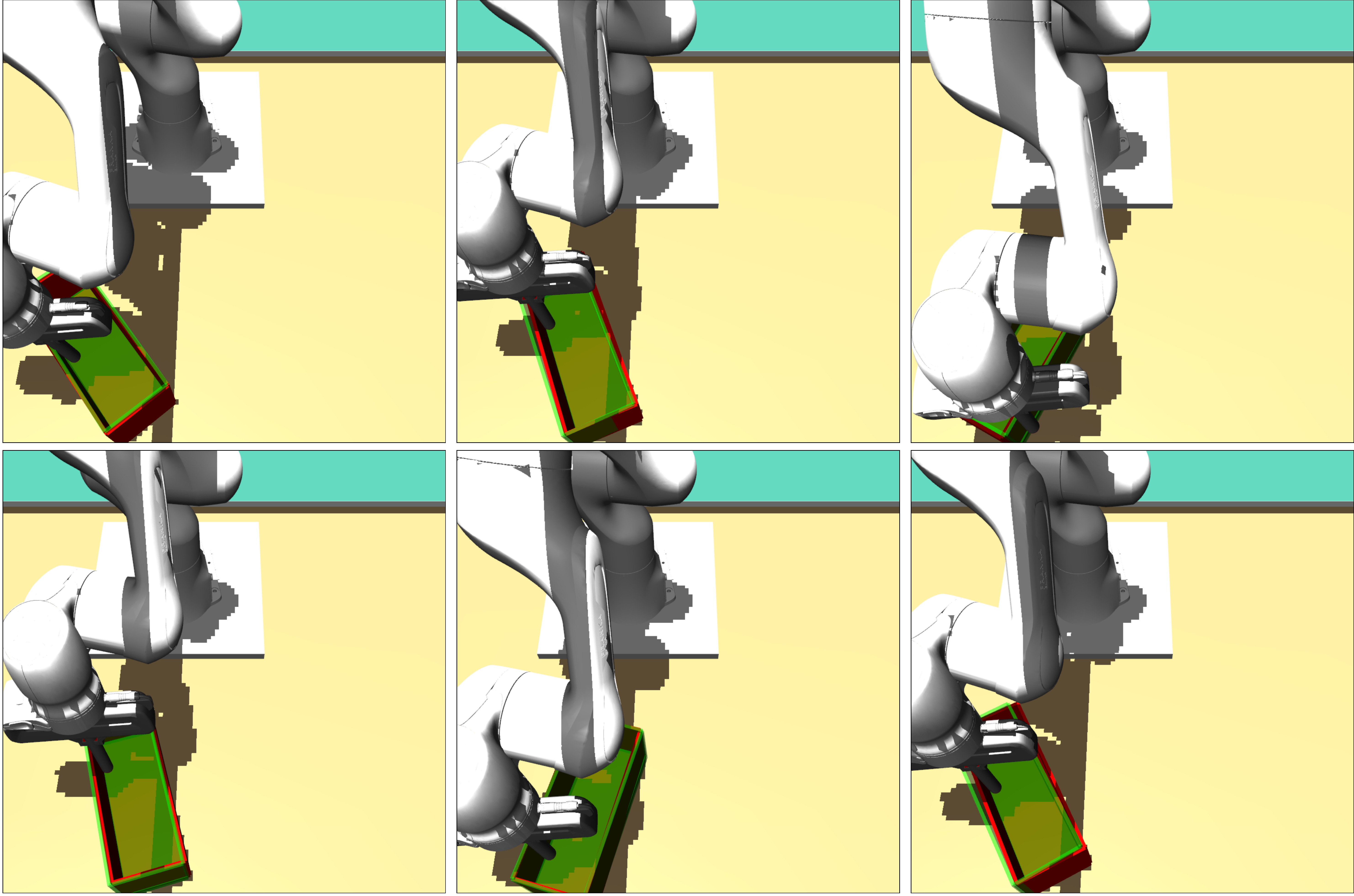}
    \caption{Visualization of the final state of the Box Pusher task on $6$ test contexts for \gls{vigor} policies trained on $24$ training contexts. The pushed boxes (red) need to align as closely as possible with the target box positions (green).}
    \label{app_fig:vis_box_pusher}
\end{figure}

\begin{figure}[t]
    \centering
    \resizebox{\textwidth}{!}{\input{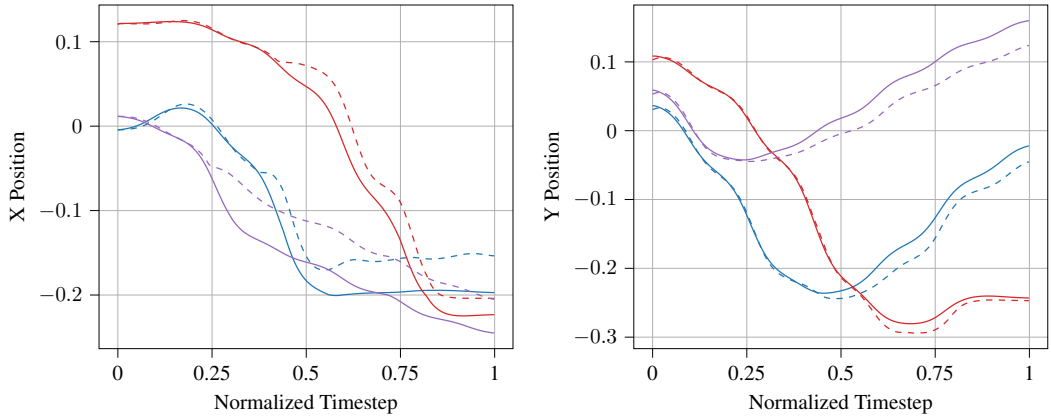}}
    \caption{Difference between planned (straight line) and executed (dotted line) trajectories for $3$ learned \gls{gmm} component means on the same test context. The executed trajectories closely match the desired ones up to the context with the box, where the force required to push the box changes the trajectory. Note that \gls{vigor} learns from geometric descriptors of planned expert demonstrations and is thus able to compensate for the contact with the box.}
    \label{app_fig:box_pusher_contacts}
\end{figure}

\paragraph{Task visualization.} Figure \ref{app_fig:vis_box_pusher} shows final states of the Box Pusher task for \gls{vigor} for $6$ test contexts on test contexts for policies trained with demonstrations from $24$ training contexts. Figure \ref{app_fig:box_pusher_contacts} shows the difference between desired and executed trajectories for $3$ learned \gls{gmm} component means on the same test context. These differences result from the contact with the (comparably heavy) box and are compensated by our approach.

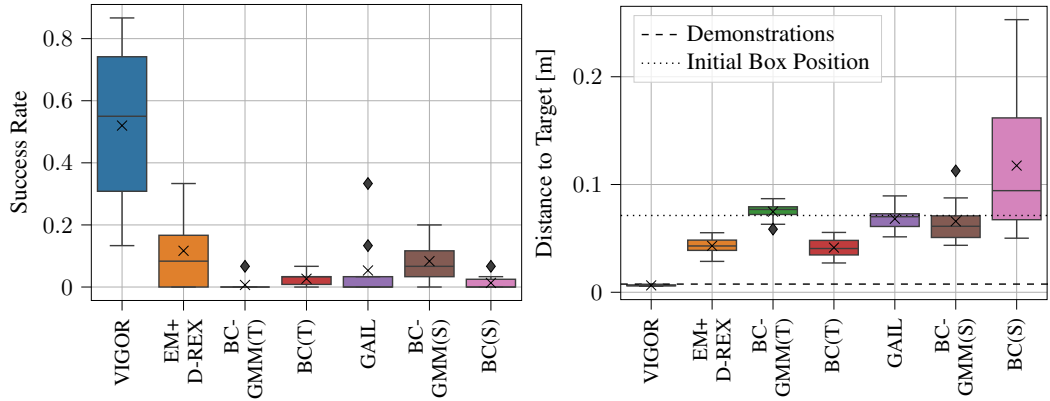
\begin{figure}[t]
    \centering
    \begin{minipage}{0.5\textwidth}
        \resizebox{\textwidth}{!}{
\begin{tikzpicture}

\definecolor{brown1926061}{RGB}{192,60,61}
\definecolor{darkgray176}{RGB}{176,176,176}
\definecolor{darkslategray61}{RGB}{61,61,61}
\definecolor{dimgray1319183}{RGB}{131,91,83}
\definecolor{forestgreen4416044}{RGB}{0,0,0}
\definecolor{mediumpurple147113178}{RGB}{147,113,178}
\definecolor{orchid213132188}{RGB}{213,132,188}
\definecolor{peru22412844}{RGB}{224,128,44}
\definecolor{seagreen5814558}{RGB}{58,145,58}
\definecolor{steelblue49115161}{RGB}{49,115,161}

\begin{axis}[
tick align=outside,
tick pos=left,
x grid style={darkgray176},
xmajorgrids,
xmin=-0.5, xmax=6.5,
xtick style={color=black},
xtick={0,1,2,3,4,5,6},
xticklabels={VIGOR,EM+\\D-REX,BC-\\GMM(T),BC(T),GAIL,BC-\\GMM(S),BC(S)},
y grid style={darkgray176},
ylabel={Success Rate},
ymajorgrids,
ymin=-0.0433333367109299, ymax=0.910000070929527,
ytick style={color=black},
height=6cm, 
width=8cm, 
x tick label style={rotate=90, font=\small, align=right}
]
\path [draw=darkslategray61, fill=steelblue49115161, semithick]
(axis cs:-0.4,0.308333337306976)
--(axis cs:0.4,0.308333337306976)
--(axis cs:0.4,0.741666659712791)
--(axis cs:-0.4,0.741666659712791)
--(axis cs:-0.4,0.308333337306976)
--cycle;
\path [draw=darkslategray61, fill=peru22412844, semithick]
(axis cs:0.6,0)
--(axis cs:1.4,0)
--(axis cs:1.4,0.16666667163372)
--(axis cs:0.6,0.16666667163372)
--(axis cs:0.6,0)
--cycle;
\path [draw=darkslategray61, fill=seagreen5814558, semithick]
(axis cs:1.6,0)
--(axis cs:2.4,0)
--(axis cs:2.4,0)
--(axis cs:1.6,0)
--(axis cs:1.6,0)
--cycle;
\path [draw=darkslategray61, fill=brown1926061, semithick]
(axis cs:2.6,0.00833333376795053)
--(axis cs:3.4,0.00833333376795053)
--(axis cs:3.4,0.0333333350718021)
--(axis cs:2.6,0.0333333350718021)
--(axis cs:2.6,0.00833333376795053)
--cycle;
\path [draw=darkslategray61, fill=mediumpurple147113178, semithick]
(axis cs:3.6,0)
--(axis cs:4.4,0)
--(axis cs:4.4,0.0333333333333333)
--(axis cs:3.6,0.0333333333333333)
--(axis cs:3.6,0)
--cycle;
\path [draw=darkslategray61, fill=dimgray1319183, semithick]
(axis cs:4.6,0.0333333333333333)
--(axis cs:5.4,0.0333333333333333)
--(axis cs:5.4,0.116666666666667)
--(axis cs:4.6,0.116666666666667)
--(axis cs:4.6,0.0333333333333333)
--cycle;
\path [draw=darkslategray61, fill=orchid213132188, semithick]
(axis cs:5.6,0)
--(axis cs:6.4,0)
--(axis cs:6.4,0.025)
--(axis cs:5.6,0.025)
--(axis cs:5.6,0)
--cycle;
\addplot [semithick, darkslategray61]
table {%
0 0.308333337306976
0 0.133333340287209
};
\addplot [semithick, darkslategray61]
table {%
0 0.741666659712791
0 0.866666734218597
};
\addplot [semithick, darkslategray61]
table {%
-0.2 0.133333340287209
0.2 0.133333340287209
};
\addplot [semithick, darkslategray61]
table {%
-0.2 0.866666734218597
0.2 0.866666734218597
};
\addplot [semithick, darkslategray61]
table {%
1 0
1 0
};
\addplot [semithick, darkslategray61]
table {%
1 0.16666667163372
1 0.333333343267441
};
\addplot [semithick, darkslategray61]
table {%
0.8 0
1.2 0
};
\addplot [semithick, darkslategray61]
table {%
0.8 0.333333343267441
1.2 0.333333343267441
};
\addplot [semithick, darkslategray61]
table {%
2 0
2 0
};
\addplot [semithick, darkslategray61]
table {%
2 0
2 0
};
\addplot [semithick, darkslategray61]
table {%
1.8 0
2.2 0
};
\addplot [semithick, darkslategray61]
table {%
1.8 0
2.2 0
};
\addplot [black, mark=diamond*, mark size=2.5, mark options={solid,fill=darkslategray61}, only marks]
table {%
2 0.0666666676600774
};
\addplot [semithick, darkslategray61]
table {%
3 0.00833333376795053
3 0
};
\addplot [semithick, darkslategray61]
table {%
3 0.0333333350718021
3 0.0666666701436043
};
\addplot [semithick, darkslategray61]
table {%
2.8 0
3.2 0
};
\addplot [semithick, darkslategray61]
table {%
2.8 0.0666666701436043
3.2 0.0666666701436043
};
\addplot [semithick, darkslategray61]
table {%
4 0
4 0
};
\addplot [semithick, darkslategray61]
table {%
4 0.0333333333333333
4 0.0333333333333333
};
\addplot [semithick, darkslategray61]
table {%
3.8 0
4.2 0
};
\addplot [semithick, darkslategray61]
table {%
3.8 0.0333333333333333
4.2 0.0333333333333333
};
\addplot [black, mark=diamond*, mark size=2.5, mark options={solid,fill=darkslategray61}, only marks]
table {%
4 0.333333333333333
4 0.133333333333333
};
\addplot [semithick, darkslategray61]
table {%
5 0.0333333333333333
5 0
};
\addplot [semithick, darkslategray61]
table {%
5 0.116666666666667
5 0.2
};
\addplot [semithick, darkslategray61]
table {%
4.8 0
5.2 0
};
\addplot [semithick, darkslategray61]
table {%
4.8 0.2
5.2 0.2
};
\addplot [semithick, darkslategray61]
table {%
6 0
6 0
};
\addplot [semithick, darkslategray61]
table {%
6 0.025
6 0.0333333333333333
};
\addplot [semithick, darkslategray61]
table {%
5.8 0
6.2 0
};
\addplot [semithick, darkslategray61]
table {%
5.8 0.0333333333333333
6.2 0.0333333333333333
};
\addplot [black, mark=diamond*, mark size=2.5, mark options={solid,fill=darkslategray61}, only marks]
table {%
6 0.0666666666666667
};
\addplot [semithick, darkslategray61]
table {%
-0.4 0.55000002682209
0.4 0.55000002682209
};
\addplot [forestgreen4416044, mark=x, mark size=3, mark options={solid,fill=black}, only marks]
table {%
0 0.520000013709068
};
\addplot [semithick, darkslategray61]
table {%
0.6 0.0833333358168602
1.4 0.0833333358168602
};
\addplot [forestgreen4416044, mark=x, mark size=3, mark options={solid,fill=black}, only marks]
table {%
1 0.116666670143604
};
\addplot [semithick, darkslategray61]
table {%
1.6 0
2.4 0
};
\addplot [forestgreen4416044, mark=x, mark size=3, mark options={solid,fill=black}, only marks]
table {%
2 0.00666666676600774
};
\addplot [semithick, darkslategray61]
table {%
2.6 0.0333333350718021
3.4 0.0333333350718021
};
\addplot [forestgreen4416044, mark=x, mark size=3, mark options={solid,fill=black}, only marks]
table {%
3 0.0266666680574417
};
\addplot [semithick, darkslategray61]
table {%
3.6 0
4.4 0
};
\addplot [forestgreen4416044, mark=x, mark size=3, mark options={solid,fill=black}, only marks]
table {%
4 0.0533333333333333
};
\addplot [semithick, darkslategray61]
table {%
4.6 0.0666666666666667
5.4 0.0666666666666667
};
\addplot [forestgreen4416044, mark=x, mark size=3, mark options={solid,fill=black}, only marks]
table {%
5 0.0833333333333333
};
\addplot [semithick, darkslategray61]
table {%
5.6 0
6.4 0
};
\addplot [forestgreen4416044, mark=x, mark size=3, mark options={solid,fill=black}, only marks]
table {%
6 0.0133333333333333
};
\end{axis}

\end{tikzpicture}}
    \end{minipage}%
    \begin{minipage}{0.5\textwidth}
        \resizebox{\textwidth}{!}{
\begin{tikzpicture}

\definecolor{brown1926061}{RGB}{192,60,61}
\definecolor{darkgray176}{RGB}{176,176,176}
\definecolor{darkslategray61}{RGB}{61,61,61}
\definecolor{dimgray1319183}{RGB}{131,91,83}
\definecolor{forestgreen4416044}{RGB}{0,0,0}
\definecolor{lightgray204}{RGB}{204,204,204}
\definecolor{mediumpurple147113178}{RGB}{147,113,178}
\definecolor{orchid213132188}{RGB}{213,132,188}
\definecolor{peru22412844}{RGB}{224,128,44}
\definecolor{seagreen5814558}{RGB}{58,145,58}
\definecolor{steelblue49115161}{RGB}{49,115,161}

\begin{axis}[
legend cell align={left},
legend style={
  fill opacity=0.8,
  draw opacity=1,
  text opacity=1,
  at={(0.03,0.97)},
  anchor=north west,
  draw=lightgray204
},
width=8cm,
height=6.0cm,
tick align=outside,
tick pos=left,
x grid style={darkgray176},
xmajorgrids,
xmin=-0.5, xmax=6.5,
xtick style={color=black},
xtick={0,1,2,3,4,5,6},
xticklabels={VIGOR,EM+\\D-REX,BC-\\GMM(T),BC(T),GAIL,BC-\\GMM(S),BC(S)},
y grid style={darkgray176},
ylabel={Distance to Target [m]},
ymajorgrids,
ymin=-0.00681775854768738, ymax=0.265355760118789,
ytick style={color=black},
x tick label style={rotate=90, font=\small, align=right}
]
\path [draw=darkslategray61, fill=steelblue49115161, semithick]
(axis cs:-0.4,0.00600801964182175)
--(axis cs:0.4,0.00600801964182175)
--(axis cs:0.4,0.00684241960310806)
--(axis cs:-0.4,0.00684241960310806)
--(axis cs:-0.4,0.00600801964182175)
--cycle;
\path [draw=darkslategray61, fill=peru22412844, semithick]
(axis cs:0.6,0.0388734013686225)
--(axis cs:1.4,0.0388734013686225)
--(axis cs:1.4,0.0484088122882202)
--(axis cs:0.6,0.0484088122882202)
--(axis cs:0.6,0.0388734013686225)
--cycle;
\path [draw=darkslategray61, fill=seagreen5814558, semithick]
(axis cs:1.6,0.0721977918740604)
--(axis cs:2.4,0.0721977918740604)
--(axis cs:2.4,0.0793373086169475)
--(axis cs:1.6,0.0793373086169475)
--(axis cs:1.6,0.0721977918740604)
--cycle;
\path [draw=darkslategray61, fill=brown1926061, semithick]
(axis cs:2.6,0.0346421213224134)
--(axis cs:3.4,0.0346421213224134)
--(axis cs:3.4,0.0481455753003506)
--(axis cs:2.6,0.0481455753003506)
--(axis cs:2.6,0.0346421213224134)
--cycle;
\path [draw=darkslategray61, fill=mediumpurple147113178, semithick]
(axis cs:3.6,0.0610500110449735)
--(axis cs:4.4,0.0610500110449735)
--(axis cs:4.4,0.072849670193428)
--(axis cs:3.6,0.072849670193428)
--(axis cs:3.6,0.0610500110449735)
--cycle;
\path [draw=darkslategray61, fill=dimgray1319183, semithick]
(axis cs:4.6,0.0508268529772601)
--(axis cs:5.4,0.0508268529772601)
--(axis cs:5.4,0.0709220410263789)
--(axis cs:4.6,0.0709220410263789)
--(axis cs:4.6,0.0508268529772601)
--cycle;
\path [draw=darkslategray61, fill=orchid213132188, semithick]
(axis cs:5.6,0.0673410692729943)
--(axis cs:6.4,0.0673410692729943)
--(axis cs:6.4,0.161810671797021)
--(axis cs:5.6,0.161810671797021)
--(axis cs:5.6,0.0673410692729943)
--cycle;
\addplot [semithick, black, dashed]
table {%
-0.5 0.00758827236363636
7.5 0.00758827236363636
};
\addlegendentry{Demonstrations}
\addplot [semithick, black, dotted]
table {%
-0.5 0.0712566969090909
7.5 0.0712566969090909
};
\addlegendentry{Initial Box Position}
\addplot [semithick, darkslategray61, forget plot]
table {%
0 0.00600801964182175
0 0.00555376502806157
};
\addplot [semithick, darkslategray61, forget plot]
table {%
0 0.00684241960310806
0 0.00746769358654683
};
\addplot [semithick, darkslategray61, forget plot]
table {%
-0.2 0.00555376502806157
0.2 0.00555376502806157
};
\addplot [semithick, darkslategray61, forget plot]
table {%
-0.2 0.00746769358654683
0.2 0.00746769358654683
};
\addplot [semithick, darkslategray61, forget plot]
table {%
1 0.0388734013686225
1 0.0286568512744982
};
\addplot [semithick, darkslategray61, forget plot]
table {%
1 0.0484088122882202
1 0.0552564983605279
};
\addplot [semithick, darkslategray61, forget plot]
table {%
0.8 0.0286568512744982
1.2 0.0286568512744982
};
\addplot [semithick, darkslategray61, forget plot]
table {%
0.8 0.0552564983605279
1.2 0.0552564983605279
};
\addplot [semithick, darkslategray61, forget plot]
table {%
2 0.0721977918740604
2 0.0630929417988149
};
\addplot [semithick, darkslategray61, forget plot]
table {%
2 0.0793373086169475
2 0.0868558790057299
};
\addplot [semithick, darkslategray61, forget plot]
table {%
1.8 0.0630929417988149
2.2 0.0630929417988149
};
\addplot [semithick, darkslategray61, forget plot]
table {%
1.8 0.0868558790057299
2.2 0.0868558790057299
};
\addplot [black, mark=diamond*, mark size=2.5, mark options={solid,fill=darkslategray61}, only marks, forget plot]
table {%
2 0.0585457856234917
};
\addplot [semithick, darkslategray61, forget plot]
table {%
3 0.0346421213224134
3 0.0271891034522534
};
\addplot [semithick, darkslategray61, forget plot]
table {%
3 0.0481455753003506
3 0.0555250001270761
};
\addplot [semithick, darkslategray61, forget plot]
table {%
2.8 0.0271891034522534
3.2 0.0271891034522534
};
\addplot [semithick, darkslategray61, forget plot]
table {%
2.8 0.0555250001270761
3.2 0.0555250001270761
};
\addplot [semithick, darkslategray61, forget plot]
table {%
4 0.0610500110449735
4 0.0514391382046232
};
\addplot [semithick, darkslategray61, forget plot]
table {%
4 0.072849670193428
4 0.0894251344925774
};
\addplot [semithick, darkslategray61, forget plot]
table {%
3.8 0.0514391382046232
4.2 0.0514391382046232
};
\addplot [semithick, darkslategray61, forget plot]
table {%
3.8 0.0894251344925774
4.2 0.0894251344925774
};
\addplot [semithick, darkslategray61, forget plot]
table {%
5 0.0508268529772601
5 0.043576197745228
};
\addplot [semithick, darkslategray61, forget plot]
table {%
5 0.0709220410263789
5 0.0875668285228522
};
\addplot [semithick, darkslategray61, forget plot]
table {%
4.8 0.043576197745228
5.2 0.043576197745228
};
\addplot [semithick, darkslategray61, forget plot]
table {%
4.8 0.0875668285228522
5.2 0.0875668285228522
};
\addplot [black, mark=diamond*, mark size=2.5, mark options={solid,fill=darkslategray61}, only marks, forget plot]
table {%
5 0.112674120548286
};
\addplot [semithick, darkslategray61, forget plot]
table {%
6 0.0673410692729943
6 0.0501852727583487
};
\addplot [semithick, darkslategray61, forget plot]
table {%
6 0.161810671797021
6 0.252984236543041
};
\addplot [semithick, darkslategray61, forget plot]
table {%
5.8 0.0501852727583487
6.2 0.0501852727583487
};
\addplot [semithick, darkslategray61, forget plot]
table {%
5.8 0.252984236543041
6.2 0.252984236543041
};
\addplot [semithick, darkslategray61, forget plot]
table {%
-0.4 0.00614589868893677
0.4 0.00614589868893677
};
\addplot [forestgreen4416044, mark=x, mark size=3, mark options={solid,fill=black}, only marks, forget plot]
table {%
0 0.00639619673263748
};
\addplot [semithick, darkslategray61, forget plot]
table {%
0.6 0.0429704610393907
1.4 0.0429704610393907
};
\addplot [forestgreen4416044, mark=x, mark size=3, mark options={solid,fill=black}, only marks, forget plot]
table {%
1 0.0431015362689155
};
\addplot [semithick, darkslategray61, forget plot]
table {%
1.6 0.0768406154641467
2.4 0.0768406154641467
};
\addplot [forestgreen4416044, mark=x, mark size=3, mark options={solid,fill=black}, only marks, forget plot]
table {%
2 0.074816726429802
};
\addplot [semithick, darkslategray61, forget plot]
table {%
2.6 0.0405964729344245
3.4 0.0405964729344245
};
\addplot [forestgreen4416044, mark=x, mark size=3, mark options={solid,fill=black}, only marks, forget plot]
table {%
3 0.0414734990810408
};
\addplot [semithick, darkslategray61, forget plot]
table {%
3.6 0.0701082128843524
4.4 0.0701082128843524
};
\addplot [forestgreen4416044, mark=x, mark size=3, mark options={solid,fill=black}, only marks, forget plot]
table {%
4 0.0681607700013574
};
\addplot [semithick, darkslategray61, forget plot]
table {%
4.6 0.061232456712799
5.4 0.061232456712799
};
\addplot [forestgreen4416044, mark=x, mark size=3, mark options={solid,fill=black}, only marks, forget plot]
table {%
5 0.0658698737768965
};
\addplot [semithick, darkslategray61, forget plot]
table {%
5.6 0.0942939780489591
6.4 0.0942939780489591
};
\addplot [forestgreen4416044, mark=x, mark size=3, mark options={solid,fill=black}, only marks, forget plot]
table {%
6 0.117558934328273
};
\end{axis}

\end{tikzpicture}}
    \end{minipage}
    \caption{
    (Left) Mean success rates on test contexts for the Box Pusher task. We find that \gls{vigor} produces the most successful demonstrations, while EM+D-REX and BC-GMM (S) are able to sometimes solve the task. 
    Interestingly, BC-GMM (T) does not produce successful solutions even though it improves the distance to target, as seen on the left of Figure \ref{fig:results_box_pusher}. We find that this corresponds to trajectories that push the box in the right general direction, but do not account for a precise combination of target positions and angle.
    (Right) Mean distance to target for training contexts of the box pushing task. Most methods reliably improve over the initial box position on training contexts. Interestingly, EM+D-REX benefits very little from evaluating on the training set, presumably because the direct correspondence to the training data is lost by the intermediate reward representation. \gls{vigor} reaches human performance on this setup, suggesting that it can match the distribution of behavioral descriptors of the human demonstrations.
    }
    \label{app_fig:train_results_box_pusher}
\end{figure}

\paragraph{Success Rates.} The left of Figure \ref{app_fig:train_results_box_pusher} shows success rates on test contexts for different approaches trained on $6$ training contexts. A demonstration is considered successful if the average distance of the corners of the final box to that of the desired final position is less than $1.5$cm, which roughly corresponds to the maximum error made by any expert demonstration.

\paragraph{Results on training contexts.} Results of \gls{vigor} trained on $6$ training contexts on the Box Pusher task for \textit{training} contexts can be seen on the right of Figure \ref{app_fig:train_results_box_pusher}.

\paragraph{Online task variants.} To further validate the effectiveness of our geometric descriptors, we experimented with variants of \gls{bc} and BC-GMM, where the distances are computed w.r.t. the current box position. 
We find that this choice of descriptors leads to diverging behavior on test contexts.

\clearpage 

\section{Ablations}
\label{app_sec:ablations}

\begin{figure}[t]
\centering
    \includegraphics[width=\textwidth]{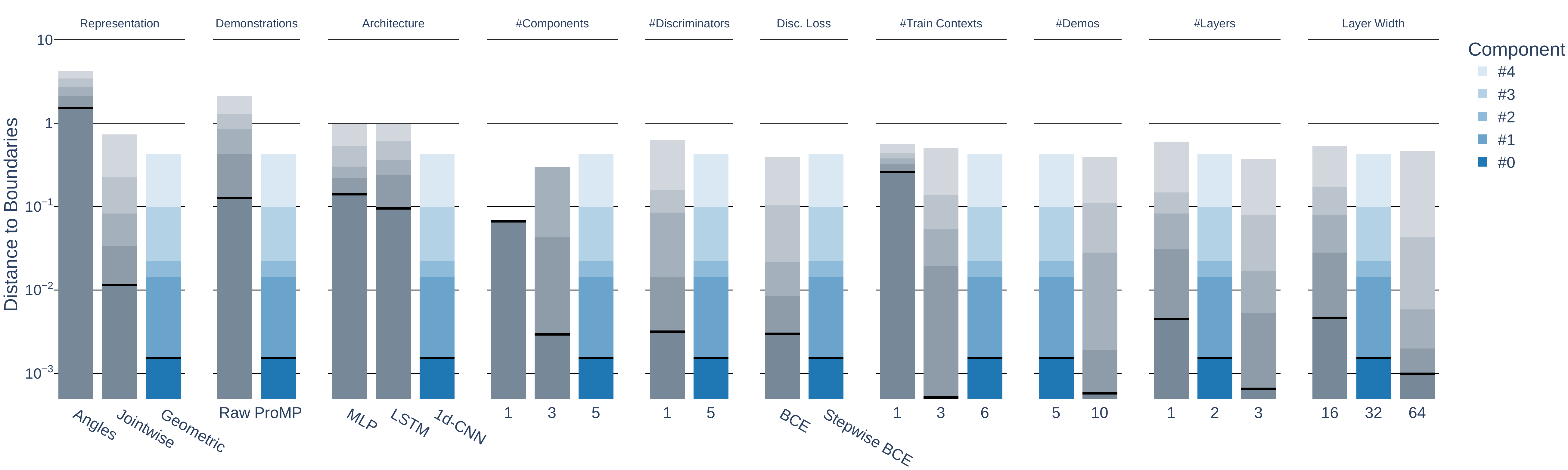}
    \caption{Full ablation study of \gls{vigor} on the Planar Reacher task. \gls{vigor} benefits from concise behavioral descriptors and is better suited to fit ProMP fits of the human demonstrations rather than the demonstrations themselves. We benefit from additional data in form of more demonstrations and more training contexts, as well as from additional components and discriminators. 
    }
    \label{app_fig:ablation_mpr_reim}
\end{figure}

\begin{figure}[t]
\centering
    \includegraphics[width=\textwidth]{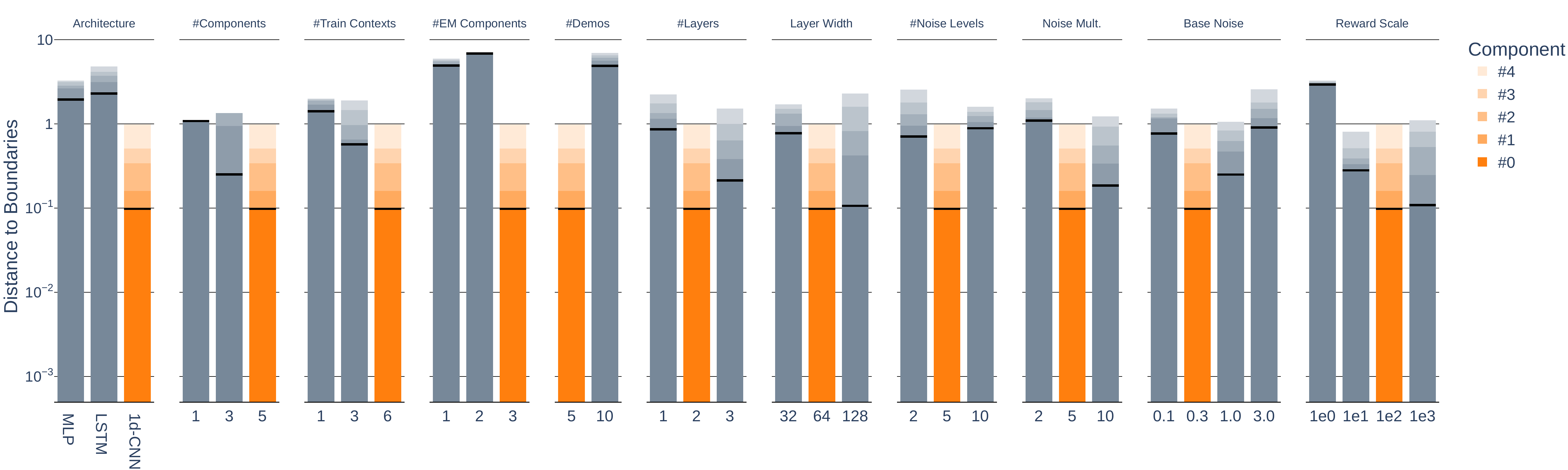}
    \caption{Ablation study of \gls{em}+\gls{drex} on the Planar Reacher task. EM+D-REX performs well if tuned correctly, but is very sensitive to its choice of hyperparameters.}
    \label{app_fig:ablation_mpr_drex}
\end{figure}

\begin{figure}[t]
\centering
    \includegraphics[width=\textwidth]{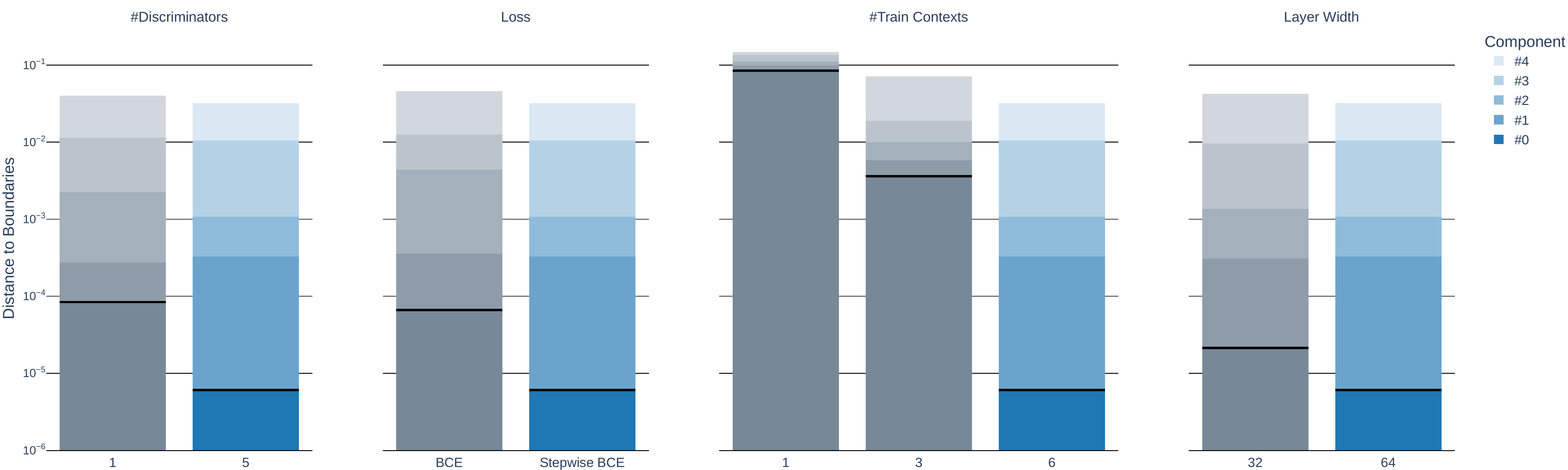}
    \caption{
    Ablation study of \gls{vigor} on the Panda Reacher task.
    }
    \label{app_fig:ablation_tel_reim}
\end{figure}

\begin{figure}[t]
\centering
    \includegraphics[width=\textwidth]{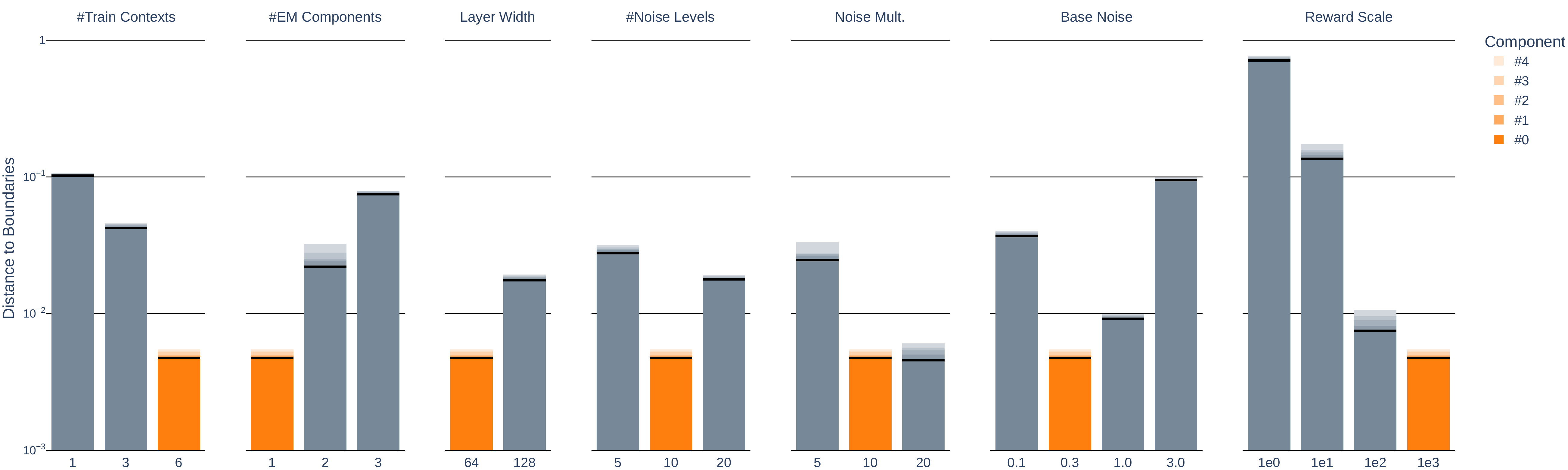}
    \caption{
    Ablation study of \gls{em}+\gls{drex} on the Panda Reacher task. 
    On this task, EM+D-REX often collapses to a single solution for the task, which can be seen by the proximity of all $5$ components in each bar. 
    The method is again very susceptible to the choice of hyperparameters, which need to be tuned on a by-task basis.
    }
    \label{app_fig:ablation_tel_drex}
\end{figure}

We perform additional ablations for \gls{vigor} and EM+D-REX on both the Planar Reacher and the Panda Reacher task. The ablations for \gls{vigor} and EM+D-REX on the Planar Reacher task can be seen in Figures \ref{app_fig:ablation_mpr_reim} and \ref{app_fig:ablation_mpr_drex} respectively. The ablations for the Panda Reacher tasks for both methods are found in Figures \ref{app_fig:ablation_tel_reim} and \ref{app_fig:ablation_tel_drex}.
All figures show the performance of all mixture components of \gls{vigor} and EM+D-REX compared to various ablations. In all figures, each bar is split into segments showing the ranked components of the learned mixture policies. In the main experiments, we evaluate the performance of the best component. This is highlighted by a black line. The blue/orange bars show the performance of \gls{vigor}/EM+D-REX respectively, the grey bars those of the ablations.

Overall, we ablate the choice of geometric descriptors, trying both a concatenation of joint angles and context (\textit{Angles}) as well as distances and velocity and acceleration for each robot joint (\textit{Jointwise}).
Additionally, we look at the performance of \gls{vigor} when trained directly on human demonstrations that are sup-sampled to $T$ time-steps (\textit{Raw}) without first fitting a ProMP on the said demonstrations.
We also evaluate the discriminator's architecture, trying out both a shared MLP per step (\textit{MLP}) with an additional encoding of the time-step, and a Long Short-Term Memory (\textit{LSTM})~\citep{hochreiter1997long}.
Both architectures are configured to be of similar size to the $1d$-CNN.
We also evaluate the effects of different numbers of mixture components (\textit{\#Components}), using an ensemble of discriminators (\textit{\#Discriminators}), the stepwise BCE loss (\textit{Disc. Loss}, c.f. Eq. \ref{eq:stepwise_bce}), and the amount of training contexts (\textit{\#Contexts}).
Finally, we inspect the number of used demonstrations per context (\textit{\#Demos)}, and the number of $1d$-CNN layers (\textit{\#Layers}) as well as the number of convolutional channels per layer (\textit{Layer Width}).

For EM+D-REX, we also look at how the number of fit EM components influences the performance, and how the algorithm-specific hyperparameters (see Appendix \ref{app_ssec:em_drex}) need to be tuned for good performance.
The ablations show that EM+D-REX performs well if tuned right, but that it is very sensitive to the choice of hyperparameters. 
The most important choice of hyperparameter seems to be the number of EM components compared to the number of demonstrations per training context. 
Depending on the task and the remaining parameters, EM+D-REX either prefers a number of components similar to the number of demonstrations, or only a single component. 
We assume that this sensitivity to the choice of hyperparameters comes from the setup of EM+D-REX. While \gls{vigor} iteratively improves its policies to match the distribution of the expert demonstrations under the geometric behavioral descriptors, EM+D-REX trains a reward function once and then uses this recovered reward function for training policies on the test contexts. 
As a result, any mistake in the recovered reward will directly influence the way that the newly trained policies are optimized.

\clearpage

\section{Hyperparameters and training details}
\label{app_sec:hyperparameters}

\subsection{Environment Parameters}
\label{app_ssec:environment_params}

Table \ref{tab:app_optuna_hyperparameters_2d} gives an overview of default environment parameters. Note that parameters with a `*' are varied in the ablations. For each environment, we randomly draw both training and test contexts at runtime from a fixed set of contexts, making sure that training and test contexts are disjoint for any given seed.

\begin{table}[ht!]
\caption{Default parameters of the different environments.}
\vspace*{0.5cm}
\centering
\label{tab:app_optuna_hyperparameters_2d}
\begin{tabular}{llll}
\toprule
Parameter & Planar Reacher & Panda Reacher & Box Pusher\\
\midrule
Trajectory length ($T$) & $30$ & $50$ & $50$\\
ProMP basis functions &  $5$ & $8$ & $7$\\
Action dimension & $5$ & $6$ & $2$\\
Context dimension & $4$ & $6$ & $3$\\
Dimension of geometric descriptors & $4$ & $4$ & $19$\\
Evaluation samples per component & $100$ & $100$ & $5$\\
Number of training contexts ($|\ctx_{\text{train}}|$) & $6$ & $6$ & $6$\\
Training Demonstrations per context & $5$ & $5$ & $3$\\
Evaluation samples per component & $100$ & $100$ & $5$\\

\bottomrule
\end{tabular}
\end{table}

\subsection{Algorithm Hyperparameters}
\label{app_ssec:algorithm_hyperparameters}

All neural networks are trained in PyTorch~\citep{pytorch2019} using the Adam~\citep{kingma2015adam} optimizer. 
We employ Dropout~\citep{srivastava2014dropout}, early stopping and Batch Normalization~\citep{ioffe2015batchnorm} to avoid overfitting. All methods are trained until convergence.
All mixture policies use $5$ components for all experiments. 

All methods use the geometric descriptors of the respective environments as input. 
We use a learning rate of $3.0e-4$ for all methods except EM+D-REX, for which we found a learning rate of $3.0e-5$ to be more stable. 
\gls{vigor} and EM+D-REX both use the equations introduced in~\citep{Becker2020Expected} for updating their policies. We found that the method performed very similarly for a range of tested KL-Bound hyperparameters, and thus set it to $0.2$ for all experiments for simplicity.
We also always draw a sufficient amount of samples to estimate a full-rank quadratic surrogate to update the parameters of their mixture models in each iteration. 
Both \gls{vigor} and EM+D-REX also make use of an ensemble with $5$ networks, Dropout of $0.2$ and use early stopping with a validation split of $0.1$. 
We do not train the categorical distribution for either \gls{vigor} or EM+D-REX. We find that Batch Normalization improves performance for EM+D-REX, but not for \gls{vigor}, and as such only use it for EM+D-REX.
Table \ref{app_table:hyperparameters_vigor} lists the remaining hyperparameters by task for \gls{vigor}, Table \ref{app_table:hyperparameters_drex} those of EM+D-REX.

\begin{table}[t]
 \caption{Hyperparameters for \gls{vigor}. Parameters with a `*' were optimized using grid searches.}
  \label{app_table:hyperparameters_vigor}
 \centering
 \begin{tabular}{llll}
 \toprule
 Parameter & Planar Reacher & Panda Reacher & Box Pusher\\
 \midrule
 $1d$-CNN layers * & $2$ & $3$ & $4$\\
 Neurons per layer * & $32$ & $64$ & $128$\\
 CNN kernel size & $5$ & $7$ & $7$\\
 Batch size & $64$ & $64$ & $64$\\
 \bottomrule
 \end{tabular}
 \end{table}

 \begin{table}[t]
 \caption{Hyperparameters for EM+D-REX. Parameters with a `*' were optimized using grid searches.}
 \label{app_table:hyperparameters_drex}
 \centering
 \begin{tabular}{llll}
 \toprule
 Parameter & Planar Reacher & Panda Reacher & Box Pusher\\
 \midrule
 $1d$-CNN layers * & $2$ & $2$ & $4$\\
 Neurons per layer * & $64$ & $64$ & $64$ \\
 CNN kernel size & $5$ & $7$ & $7$\\
 Batch size & $128$ & $128$ & $128$\\
 Fit EM Components * & $3$ & $1$ & $1$\\
 Total EM Samples & $8192$ & $16384$ & $16384$\\
 Base Noise * & $0.3$ & $0.3$ & $0.3$\\
 Noise Levels * & $5$ & $10$ & $5$\\
 Maximum Noise Multiplier * & $5$ & $10$ & $2$\\
 Reward Scale * & $100$ & $1000$ & $1000$ \\
 \bottomrule
 \end{tabular}
 \end{table}

For both BC and BC-GMM we find that training for $3000$ and $30000$ epochs on the state-action and trajectory-based settings works best respectively. Both use an entropy regularization term with a weight of $1.0e-3$. 
For BC-GMM, we additionally add an option to either train or not train the categorical distribution of the mixture to our hyperparameter search. 
If not trained, the distribution defaults to a uniform distribution over all components, similar to that of \gls{vigor} and EM+D-REX.
Remaining hyperparameters by task for BC(S) and BC-GMM(S) are given in Tables \ref{app_table:hyperparameter_bc_s} and \ref{app_table:hyperparameters_mbc_s} respectively. Those of BC(T) and BC-GMM(T) are given in Tables \ref{app_table:hyperparameters_bc_e} and \ref{app_table:hyperparameters_mbc_e}.
Hyperparameters for \gls{gail} are listed in Table \ref{app_table:hyperparameters_gail}.

 \begin{table}[t]
 \caption{Hyperparameters for state-action-based Behavioral Cloning (BC(S)). Parameters with a `*' were optimized using grid searches.}
 \label{app_table:hyperparameter_bc_s}
 \centering
 \begin{tabular}{llll}
 \toprule
 Parameter & Planar Reacher & Panda Reacher & Box Pusher \\
 \midrule
 MLP layers * & $2$ & $2$ & $3$\\
 Neurons per layer * & $64$ & $32$ & $64$\\
 Batch size * & $128$ & $64$ & $64$\\
 Include target encoding & False & True & --\\
 State-action framestacks * & $1$  & $1$ & $1$\\
 \bottomrule
 \end{tabular}
 \end{table}
 
 \begin{table}[ht!]
 \caption{Hyperparameters for state-action-based Behavioral Cloning with a Gaussian Mixture Model policy (BC-GMM(S)). Parameters with a `*' were optimized using grid searches.}
  \label{app_table:hyperparameters_mbc_s}
 \centering
 \begin{tabular}{llll}
 \toprule
 Parameter & Planar Reacher & Panda Reacher & Box Pusher\\
 \midrule
 MLP layers * & $2$ & $2$ & $3$\\
 Neurons per layer * & $64$ & $64$ & $64$\\
 Batch size * & $64$ & $64$ & $64$ \\
 Include target encoding & False & True & --\\
 State-action framestacks * & $5$ & $1$ & $1$\\
 Train categorical distribution * & True & True & False\\
 \bottomrule
 \end{tabular}
 \end{table}

 \begin{table}[t]
 \caption{Hyperparameters for trajectory-based Behavioral Cloning (BC(T)). Parameters with a `*' were optimized using grid searches.}
 \label{app_table:hyperparameters_bc_e}
\centering
\begin{tabular}{llll}
  \toprule
 Parameter & Planar Reacher & Panda Reacher & Box Pusher\\
 \midrule
 MLP layers * & $4$ & $4$ & $3$ \\
 Neurons per layer * & $256$ & $64$ & $128$\\
 Batch size * & $4$ & $4$ & $4$\\
 \bottomrule
 \end{tabular}
 \end{table}

 \begin{table}[t]
 \caption{Hyperparameters for trajectory-based Behavioral Cloning with a Gaussian Mixture Model policy (BC-GMM(T)). Parameters with a `*' were optimized using grid searches.}
 \label{app_table:hyperparameters_mbc_e}
 \centering
 \begin{tabular}{llll}
 \toprule
 Parameter & Planar Reacher & Panda Reacher & Box Pusher\\
 \midrule
 MLP layers * & $4$ & $3$ & $3$ \\
 Neurons per layer * & $64$ & $64$ & $128$ \\
 Batch size & $4$ & $4$ & $4$\\
 Train categorical distribution * & False & False & False\\
 \bottomrule
 \end{tabular}
 \end{table}

 \begin{table}[ht!]
 \caption{Hyperparameters for GAIL. Parameters with a `*' were optimized using grid searches.}
  \label{app_table:hyperparameters_gail}
 \centering
 \begin{tabular}{llll}
 \toprule
 Parameter & Planar Reacher & Panda Reacher & Box Pusher\\
 \midrule
 Discriminator MLP layers * & $2$ & $2$ & $3$ \\
 Discriminator neurons per layer * & $32$ & $64$ & $64$\\
 Policy MLP layers & $2$ & $2$ & $2$\\
 Policy MLP neurons per layer & $32$ & $32$ & $32$\\
 Batch size (Discriminator) * & $128$ & $64$ & $64$\\
 Batch size (Policy) * & $128$ & $64$ & $256$ \\
 Total time-steps * & $1e6$ & $3e6$ & $1e6$ \\
 Include target encoding & True & False & --\\
 State-action framestacks * & $5$ & $5$ & $1$ \\
 Share networks & False & False & False\\
 Policy steps & $2048$ & $2048$ & $2048$\\
 \bottomrule
 \end{tabular}
 \end{table}

\end{document}